\documentclass{article} 

\usepackage{iclr2027_conference,times}

\usepackage{amsmath,amsfonts,bm}

\def\eqref#1{equation~\ref{#1}}

\def\1{\bm{1}}

\DeclareMathAlphabet{\mathsfit}{\encodingdefault}{\sfdefault}{m}{sl}
\SetMathAlphabet{\mathsfit}{bold}{\encodingdefault}{\sfdefault}{bx}{n}

\usepackage{hyperref}
\usepackage{url}
\usepackage[table]{xcolor}

\usepackage{graphicx}
\usepackage{caption}
\usepackage{booktabs}
\usepackage{multirow}
\usepackage{array}
\usepackage{pifont}
\usepackage{wrapfig}
\usepackage{amssymb}
\usepackage{tabularx}
\definecolor{evalgray}{gray}{0.93}
\definecolor{overlapred}{RGB}{180,45,45}
\newcommand{\overlap}[1]{\textcolor{overlapred}{#1}}
\newcommand{\zeromark}{\textsuperscript{\textcolor{red}{$\blacklozenge$}}}
\newcommand{\fewmark}{\textsuperscript{\textcolor{green!60!black}{$\blacktriangle$}}}

\title{SegBanana: Steering Unified Multimodal Models into Medical Segmenters}

\author{Xiaoye Liang$^{1,3}$ \quad
Ye Yan$^{1}$ \quad
Mingze Yin$^{2}$ \quad
Shikun Feng$^{3}$ \quad
Mai Xu$^{1}$ \quad
Haiguang Liu$^{3}$ \\
\textbf{Lai Jiang}$^{1}$\thanks{Corresponding author} \quad
\textbf{Yiheng Zhu}$^{3*}$\\
$^{1}$Beihang University \\
$^{2}$Zhejiang University \\
$^{3}$Zhongguancun Academy \\
\texttt{jianglai.china@buaa.edu.cn, zhuyiheng@zgci.ac.cn}
}

\iclrfinalcopy 
\begin{document}

\maketitle

\begin{abstract}
Medical image segmentation remains challenging in practical deployment, as models often struggle to generalize beyond the distributions covered by their training data and high-quality pixel-level annotations are typically unavailable for adaptation.
Inspired by the cross-task transferability of large language models, we investigate whether unified multimodal models (UMMs) can transfer their pretrained visual understanding, reasoning, and generation capabilities to medical image segmentation without task-specific post-training. By recasting segmentation as structured visual generation, we find that frontier UMMs (e.g., Nano Banana) already exhibit basic segmentation capabilities across diverse clinical scenarios, but still struggle with challenging tasks requiring specialized anatomical or domain-specific knowledge. We further show that these limitations can be effectively mitigated by incorporating visual anatomical knowledge from in-context exemplars, expanding candidate solutions through repeated sampling, and refining suboptimal predictions via targeted editing.
Motivated by these observations, we propose SegBanana, to our knowledge, the first agentic visual generation framework for training-free medical image segmentation. 
SegBanana builds on a frozen UMM as the core generative model, augmented with Anatomy-Aware Knowledge Retrieval and Comparative Quality Critique to unlock its potential segmentation capability. A State-Aware Multimodal Controller maintains structured state and iteratively orchestrates these tools, repeatedly refining intermediate predictions toward higher-quality masks.
Across eight medical segmentation datasets, SegBanana achieves an average mDice of $77.45\%$, outperforming representative generalist (SAM3 and SegGPT) and medical-specific (BiomedParse and MedSAM3) baselines by at least $14.93$ points, while remaining robust to out-of-domain visual supports. Ablations further confirm the effectiveness of agentic inference and the contributions of retrieval, critique, and structured state.
\end{abstract}

\section{Introduction}
Medical image segmentation remains difficult to deploy reliably in new clinical scenarios.
Changes in imaging devices, acquisition protocols, clinical sites, patient populations, or disease presentations can induce substantial distribution shifts, under which most training-based models generalize poorly beyond their training distributions \citep{niu2024survey}.
Meanwhile, the pixel-level annotations required for adaptation are often scarce and costly to acquire at scale.
These limitations motivate more flexible paradigms that can generalize to new clinical scenarios with limited task-specific supervision.

\begin{figure}[tb]
  \centering
  \includegraphics[width=1\textwidth]{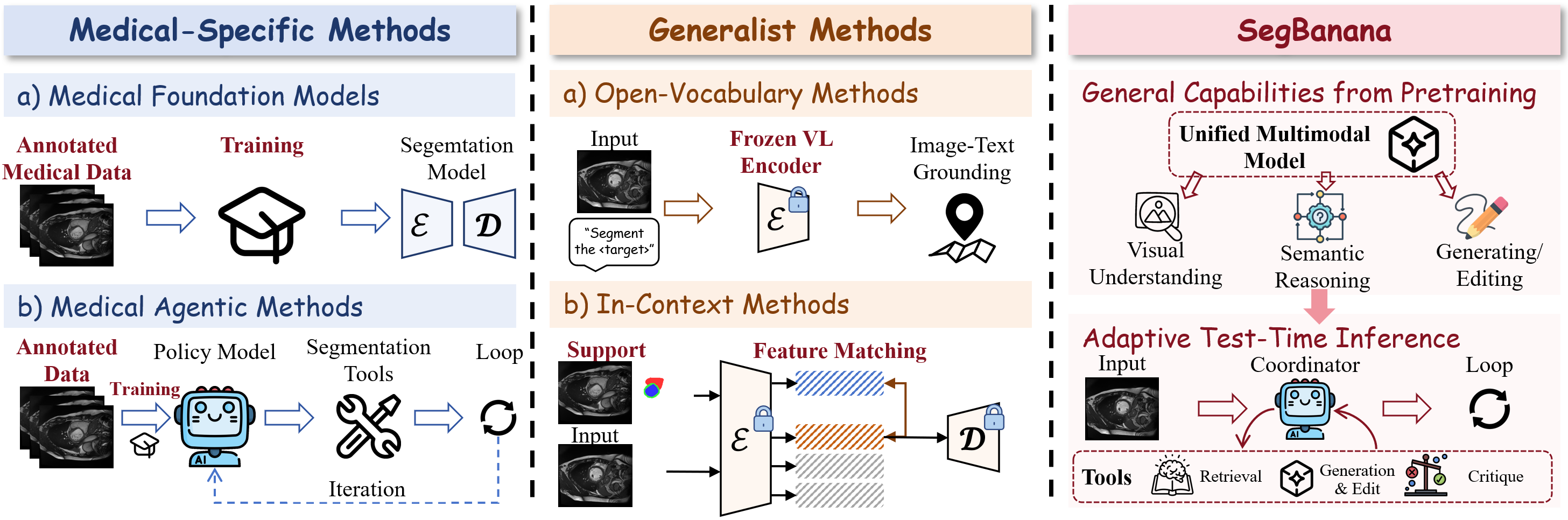}

  \caption{Comparison of generalization paradigms for medical image segmentation, including medical-specific methods, generalist methods, and our SegBanana.
  }
  \label{fig:teaser}

\end{figure}

Recent efforts toward this objective have explored multiple directions, including medical-specific and generalist methods, as illustrated in Fig.~\ref{fig:teaser}.
Medical-specific foundation models \citep{MedSAM,zhu2024medical,biomedparse} and recent agentic methods \citep{jiang2026ibisagent,huang2026medsegagent} build on capabilities acquired from medical training data or pretrained segmentation tools, and often degrade on tasks or domains beyond this coverage.
Generalist methods instead leverage broadly pretrained representations or annotated supports at inference time, but open-vocabulary methods \citep{barsellotti2025talking,zeng2024maskclip++,carion2026sam} may lack specialized anatomical knowledge, while in-context methods \citep{cuttano2026insid3,zhang2024bridge,zhang2305personalize} rely heavily on well-matched supports and degrade under domain shift.
Overall, existing methods remain tied to capabilities acquired during training or direct correspondence with annotated supports, motivating more flexible  transfer paradigms.

Recent advances in large language models have shown that large-scale generative pretraining enables broad task transfer without task-specific updates \citep{brown2020language,chowdhery2023palm}. Inspired by this, we ask: \textit{can unified multimodal models (UMMs), which are similarly pretrained at scale with visual understanding, reasoning, and image generation capabilities, transfer to diverse medical segmentation tasks?} Unlike conventional segmentation models that rely on task-specific prediction spaces, UMMs define segmentation tasks through a shared generative interface conditioned on instructions and visual context, potentially allowing pretrained capabilities to be recomposed for new tasks at test time. Recent studies on natural images, including PixelArena \citep{liang2025pixelarena} and VisionBanana \citep{gabeur2026image}, provide encouraging evidence that frontier UMMs already possess basic zero-shot segmentation capability and can be further extended to broader vision tasks through instruction tuning. However, how to effectively unlock this intrinsic capability for medical image segmentation remains unclear.

Our study reveals that UMMs already exhibit basic zero-shot medical segmentation capability, particularly on tasks with salient global shape or appearance cues. However, their performance degrades substantially on anatomically demanding tasks that require fine-grained discrimination of adjacent or nested structures, indicating that generic visual knowledge alone is insufficient when accurate segmentation depends on specialized anatomical and structural understanding.
We further find that visual anatomical knowledge from in-context exemplars can compensate for insufficient task-specific priors, repeated sampling and diverse conditioning can expand the accessible prediction space, and targeted editing can effectively refine suboptimal predictions when sufficiently informative correction guidance is available.

Motivated by these observations, we propose SegBanana, a training-free medical segmentation agent that harnesses a frozen UMM through stateful, feedback-driven tool use.
By casting medical image segmentation as structured visual generation, SegBanana uses a frozen UMM to generate and refine segmentation masks, augmented by Anatomy-Aware Knowledge Retrieval to supplement missing task-specific knowledge and Comparative Quality Critique to evaluate the quality of current masks and diagnose residual errors. A State-Aware Multimodal Controller reasons over the structured interaction trajectory and tool observations, and adaptively orchestrates retrieval, generation, editing, and critique to iteratively improve the segmentation result.
Our contributions are three-fold:
\begin{itemize}
    \item We systematically study the intrinsic medical segmentation capability of UMMs and identify knowledge augmentation, output exploration, and targeted editing as effective mechanisms for training-free transfer.

    \item We introduce SegBanana, a training-free medical segmentation agent that harnesses a frozen UMM through stateful, feedback-driven tool use, dynamically orchestrating knowledge retrieval, visual generation and editing, and comparative critique.
    
    \item Experiments on eight datasets show that SegBanana substantially improves frozen UMMs, achieves strong generalization across diverse clinical scenarios, and remains robust when only out-of-domain visual supports are available.
\end{itemize}

\section{Related Work}
\textbf{Medical Image Segmentation Models.} 
Existing medical segmentation models \citep{MedSAM,zhu2024medical,biomedparse,wang2025sam,magg2024zeroshotcapabilitysamfamilymodels} achieve broad coverage but still rely on task-specific training, which limits adaptation to unseen domains. 
Recent agentic medical segmentation frameworks \citep{jiang2026ibisagent,huang2026medsegagent} enable iterative segmentation, but depend on trained policy models and segmentation tools, limited by their training distributions. 
General open-vocabulary segmentation models \citep{barsellotti2025talking,zeng2024maskclip++,carion2026sam} improve flexibility, but their performance is often constrained by limited medical knowledge and dependence on user-provided prompts. In-context segmentation methods \citep{cuttano2026insid3,zhang2305personalize,zhang2024bridge,wang2023images,wang2023seggpt} further adapt to new tasks through visual demonstrations, yet largely rely on feature correspondence and remain sensitive to the provided context. Instead, we explore an understanding-driven transfer paradigm that reasons over task semantics and visual context for training-free adaptation to unseen clinical scenarios.

\textbf{Unified Multimodal Models.}
Unified multimodal models (UMMs) integrate visual understanding and generation within a framework \citep{team2024chameleon,xie2025show,wang2024emu3,wu2025janus,deng2025bagel,comanici2025gemini,lin2025uniworld}. 
Chameleon \citep{team2024chameleon}, Show-o \citep{xie2025show}, and Emu3 \citep{wang2024emu3} unify multimodal modeling through shared token prediction, while Janus \citep{wu2025janus} decouples visual representations for understanding and generation within a unified backbone. 
Later models, including Nano Banana, extend these capabilities toward multimodal reasoning and instruction-guided image editing \citep{deng2025bagel,comanici2025gemini,lin2025uniworld}. 
Building on pretrained UMMs, a growing line of work adapts them to general-purpose vision tasks through task-level supervision and post-training \citep{gabeur2026image,han2026vision}. 
In contrast, we investigate whether the intrinsic visual capabilities of pretrained UMMs can be unlocked for medical image segmentation through agentic inference alone.

\textbf{Agentic Frameworks for Image Generation and Editing.}
Agentic frameworks for image generation \citep{venkatesh2025crea,nabati2024preference,wang2025imagent,jiang2026genagent,shen2026agentic} enhance generative models through multi-agent collaboration, preference-aware interaction, self-evaluation, and iterative tool use, enabling more diverse generation, improved user alignment, and finer-grained control over generated content.
Agentic image editing frameworks \citep{zeng2026mira,zhang2026intentedit,zhao2026imageedit} further introduce iterative reasoning, collaborative planning and reflection, and reinforcement-learned coordination to improve edit quality, controllability, and alignment with complex instructions.
These methods primarily use agentic inference to improve generation or editing within their original task settings. In contrast, we investigate whether agentic inference can transfer the capabilities of pretrained UMMs beyond generation and editing to training-free medical image segmentation.


\section{Observation}
\label{sec:observations}
In this section, we characterize the intrinsic segmentation behavior of UMMs. Viewing segmentation as conditional generation, we study how external conditioning and inference-time sampling strategies shape the accessible output space. Moreover, leveraging the inherent editing capability of these models, we investigate whether existing predictions can be further revised and improved. 

\textbf{Generation Capability:}
\textit{UMMs exhibit basic medical segmentation capability, but remain limited on tasks requiring specialized anatomical knowledge.}
We evaluate UMMs on three representative tasks spanning different modalities, anatomical targets, scales, and spatial structures, as summarized in Tab.~\ref{tab:task_characteristics}. For each task, the model receives only the input image and a task-specific instruction, e.g., ``Segment the \texttt{<target>}'', and directly produces a segmentation mask in a zero-shot manner. Then, the generated image is converted into a standardized discrete mask through pixel-wise color clustering. As shown in Tab.~\ref{tab:task_characteristics}, UMMs perform substantially better on tasks with salient global shape or appearance cues, such as ISIC, than on ACDC and Drishti-GS, which require finer discrimination of adjacent or nested anatomical structures. These results suggest that UMMs can leverage generic visual cues for medical segmentation, but remain limited when accurate prediction requires specialized anatomical and structural knowledge.

\textbf{Generation Conditioning:}
\textit{Relevant visual supports effectively compensate for insufficient task-specific knowledge.}
We provide image--mask exemplars as visual supports to supplement the UMMs with task-relevant anatomical knowledge. To examine how support selection affects segmentation, we compare randomly selected exemplars with query-relevant supports retrieved using DINO features \citep{simeoni2025dinov3}. As shown in Tab.~\ref{tab:task_characteristics}, query-relevant supports yield consistent improvements on tasks requiring specialized anatomical knowledge, whereas random supports provide less reliable gains and may introduce irrelevant context.

\begin{table*}[t]
\centering
\caption{
Task characteristics and performance under text-only and support-augmented prompting, where Random samples a support and Retrieved selects a query-relevant one.
}
\vspace{-0.5em}
\label{tab:task_characteristics}
\resizebox{1\textwidth}{!}{
\begin{tabular}{@{}l|ccccc|ccc@{}}
\toprule
\multirow{2}{*}{\textbf{Dataset}}& \multicolumn{5}{c|}{\textbf{Task Characteristics}}
& \multicolumn{3}{c}{\textbf{mDice (\%)}} \\
\cmidrule(lr){2-6}
\cmidrule(lr){7-9}
& \textbf{Modality}
& \textbf{Target}
& \textbf{Size}
& \textbf{Structure}
& \textbf{Main Challenge}
& \textbf{Text-only}
& \textbf{Random}
& \textbf{Retrieved} \\
\midrule

ACDC \citep{bernard2018deep}
& MRI
& RV / MYO / LV
& Small--Med.
& Adjacent
& Anatomical coupling
& 36.89
& 31.79
& \textbf{45.30} \\

Drishti-GS \citep{sivaswamy2014drishti}
& Fundus
& Disc / Cup
& Small
& Nested
& Subtle boundaries
& 43.73
& 53.57
& \textbf{57.49} \\

ISIC \citep{codella2019skin}
& Dermoscopy
& Lesion
& Med.--Large
& Irregular
& Appearance variation
& 74.68
& 79.62
& \textbf{80.65} \\


\bottomrule
\end{tabular}
}
\end{table*}
\begin{figure}[tb]
  \centering
  \vspace{-0.5em}
  \includegraphics[width=.9\textwidth]{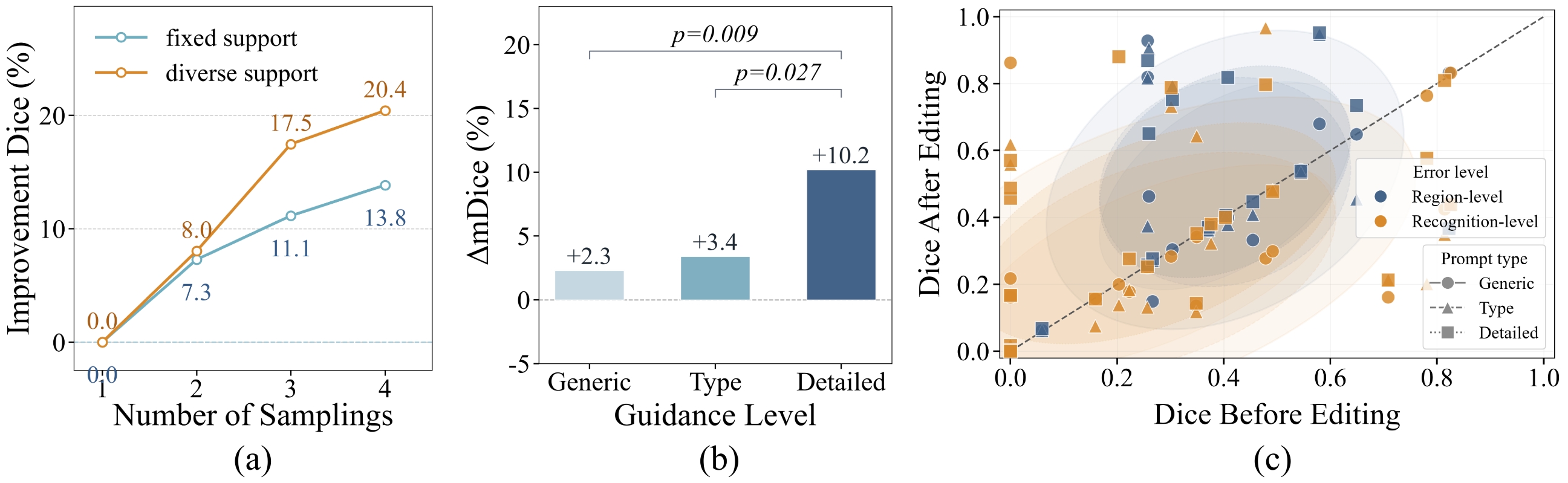}
  \vspace{-1.1em}
  \caption{Inference-time exploration and refinement.
(a) Best-of-\(N\) performance with increasing sampling.
(b) Improvement under varying editing guidance.
(c) Before--after Dice across error levels and prompt types, with shaded regions indicating their distributions.
  }
  \vspace{-1.5em}
  \label{fig:obs2}
\end{figure}
\textbf{Generation Exploration:}
\textit{Repeated sampling reveals diverse predictions, while varying the conditioning further expands the accessible output space.}
We first examine whether stochastic generation can expose better segmentation predictions beyond a single output. On ACDC, we sample four outputs under the same query, instruction, and visual support, and evaluate oracle Best-of-$N$ performance. As shown in Fig.~\ref{fig:obs2}(a), Best-of-$N$ steadily improves as the sampling budget increases, indicating that even fixed conditioning yields substantial output diversity and that a single generation may overlook better predictions. We then vary the retrieved support while keeping the generation budget fixed. As shown in Fig.~\ref{fig:obs2}(a), sampling across different supports consistently outperforms repeated sampling with a fixed support, suggesting that conditioning diversity further broadens the accessible prediction space and exposes higher-quality candidate predictions.

\textbf{Editing Behavior:}
\textit{Detailed correction guidance substantially improves refinement, while editing is most reliable for region-level errors.}
Beyond generating alternative predictions, UMMs can refine existing segmentations through editing. We study this capability on ACDC using the query image, a prediction, and a correction instruction. To examine the effect of instruction specificity, we compare three levels of guidance under the same image and mask: \emph{generic editing}, which only requests correction; \emph{type-guided editing}, which specifies the error type; and \emph{detailed editing}, which localizes the error and describes the desired revision. As shown in Fig.~\ref{fig:obs2}(b), more specific guidance yields larger improvements, indicating that effective refinement benefits from precise and actionable instructions.
We further analyze editing effectiveness across error types. Region-level errors preserve target identity and overall structure, whereas recognition-level errors involve incorrect identification or localization. As shown in Fig.~\ref{fig:obs2}(c), editing is more reliable for region-level errors, suggesting that it is better suited to refining plausible predictions than recovering fundamentally incorrect ones.

Further analyses of these behaviors are provided in Sec.~\ref{sec:obs_supp}, including diversity-aware retrieval and the complementary effects of editing and resampling. Together, these observations motivate an agentic controller that can dynamically decide when to acquire knowledge, explore new masks, or refine an existing prediction according to the evolving segmentation state.

\section{Method}

To unlock the potential segmentation capability of pretrained UMMs, we cast medical image segmentation as structured visual generation and augment a frozen UMM with external anatomical knowledge and comparative feedback under state-aware agentic control, enabling iterative exploration and refinement of segmentation predictions.

\begin{figure}[tb]
  \centering
  \includegraphics[width=1\textwidth]{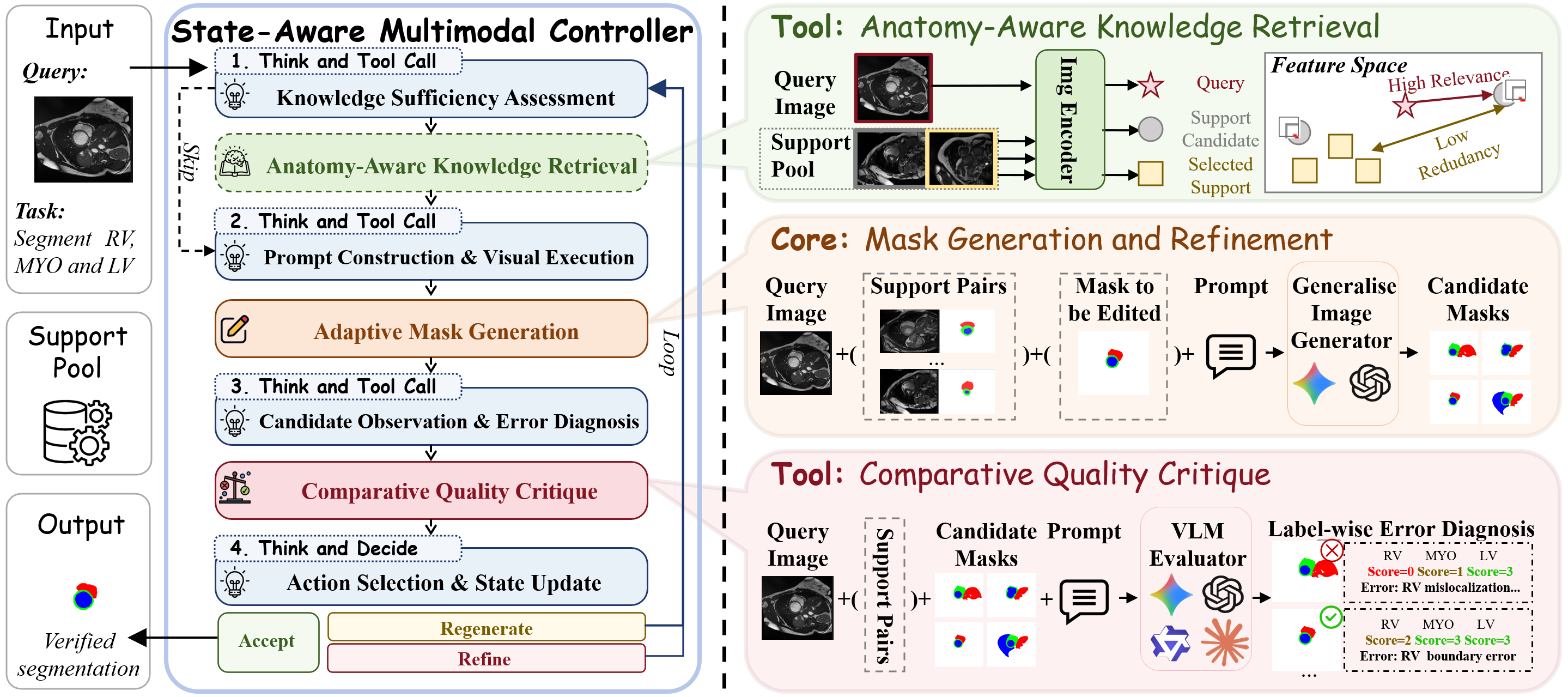}
  \vspace{-2em}
  \caption{
  Overview of SegBanana. A State-Aware Multimodal Controller orchestrates Anatomy-Aware Knowledge Retrieval, UMM-based Mask Generation and Refinement, and Comparative Quality Critique, using state and feedback to adapt inference.
  }
  \label{fig:framework}
\end{figure}

\textbf{Method Overview.}
We present an overview of SegBanana in Fig.~\ref{fig:framework}.
Given a query image, a segmentation instruction, and an external pool of annotated visual supports, SegBanana performs segmentation through closed-loop visual generation and refinement centered on a frozen UMM.
At each iteration, a State-Aware Multimodal Controller reasons over the current state and observations to select the next tool action.
When additional task-specific anatomical knowledge is required, it invokes
Anatomy-Aware Knowledge Retrieval to acquire relevant visual support.
The controller then synthesizes an action-specific instruction and invokes the
frozen UMM to either explore a new segmentation mask or edit the retained prediction.
The resulting candidates are then examined by Comparative Quality Critique,  whose preference and residual-error diagnosis are returned as feedback.
This feedback updates the structured state and closes the loop, allowing the
controller to reassess the current prediction and adapt its next action until
Accept is selected or the inference budget is exhausted.

\subsection{Capability-Enhancing Segmentation Toolset}
\label{sec:method_tools}

\textbf{Anatomy-Aware Knowledge Retrieval.}
Given a query image $I$, task instruction $T$, and support pool
$\mathcal{D}_s=\{(I_i,M_i)\}_{i=1}^{N}$,
we adopt maximal marginal relevance (MMR) \citep{mmr} to balance query relevance and support redundancy:
\begin{equation}
i_k=
\arg\max_{i\notin \mathcal{S}_{k-1}}
\left[
\lambda\, s_q(i)
-(1-\lambda)\max_{j\in\mathcal{S}_{k-1}} s(i,j)
\right],
\end{equation}
where $s_q(i)$ measures query--support similarity and $s(i,j)$ measures similarity between support examples.
After $K$ selections, the retriever returns
\begin{equation}
\mathcal{S}_t
=
\mathcal{R}(I,T,\mathcal{D}_s)
=
\{(I_{i_k},M_{i_k})\}_{k=1}^{K},
\end{equation}
which provides task-relevant anatomical, semantic, and appearance cues for subsequent generation.

\textbf{Mask Generation and Refinement.}
Given a query image $I$ and task instruction $T$, the controller invokes the frozen UMM through two visual actions:
\begin{equation}
M_t =
\begin{cases}
G(I,T,\mathcal{K}_t), & \textsc{Generate},\\
G(I,B_t,E_t), & \textsc{Edit},
\end{cases}
\end{equation}
where $\mathcal{K}_t$ denotes the context, which may include retrieved image--mask supports, $B_t$ is the retained best prediction, and $E_t$ is a targeted correction instruction synthesized from the current state.
The two modes enable exploration of alternative predictions and local refinement of promising results. 

\textbf{Comparative Quality Critique.}
We employ a frozen multimodal VLM as a feedback tool that comparatively
evaluates competing predictions and diagnoses residual errors.
Let $\mathcal{M}_t=\{M_t^{(1)},\ldots,M_t^{(n)}\}$ denote the newly produced candidates.
If no retained prediction is available, the candidates are compared sequentially to determine a provisional champion.
Otherwise, the retained best prediction $B_t$ is introduced as a baseline and challenged by the new candidates:
\begin{equation}
(\hat{M}_t,O_t^{\mathrm{crit}})
=
\operatorname{Tournament}_{\mathcal{V}}
\left(
I,\,
\mathcal{M}_t,\,
B_t
\right),
\end{equation}
where $\hat{M}_t$ denotes the tournament winner and $O_t^{\mathrm{crit}}$ contains the corresponding visual evaluation and residual-error diagnosis.
The critic compares predictions based on target identity, location, shape, topology, and boundary alignment.
For multi-label tasks, different labels may be retained from different candidates and composed into the current prediction.
The resulting preference and error diagnosis are returned as structured
feedback, forming the observation that drives the controller's next action.

\subsection{State-Aware Multimodal Controller}
\label{sec:method_controller}

SegBanana employs a frozen multimodal VLM as a ReAct-style controller that reasons over the query, task instruction, stage-relevant observations, and compact interaction trajectory to determine the next tool action and refinement strategy.

\textbf{Structured State.}
Rather than replaying the full interaction trajectory, SegBanana maintains a compact structured state
\begin{equation}
\mathcal{Z}_t=\{\mathcal{C}_t,\mathcal{H}_t\},
\end{equation}
comprising a \emph{Current State} $\mathcal{C}_t$ and a \emph{Historical State} $\mathcal{H}_t$.
The Current State summarizes the observations required for the next action decision, including the latest prediction, visual evaluation, residual-error diagnosis.
The Historical State summarizes cross-iteration information:
\begin{equation}
\mathcal{H}_t =
\left\{
B_t,\,
\mathcal{E}_t,\,
\mathcal{H}^{\mathrm{edit}}_t,\,
\mathcal{H}^{\mathrm{ret}}_t
\right\},
\end{equation}
where $B_t$ denotes the retained best prediction,
$\mathcal{E}_t$ summarizes accumulated error observations,
$\mathcal{H}^{\mathrm{edit}}_t$ tracks previous editing attempts and outcomes,
and $\mathcal{H}^{\mathrm{ret}}_t$ tracks explored supports and their observed utility.
This compact state serves as the agent's working memory, preserving
cross-iteration experience while exposing only stage-relevant information
to each module, enabling the controller to track unresolved errors and avoid
ineffective actions without replaying the full interaction trajectory.


\textbf{State-Aware Tool Use.}
At each iteration, the controller first determines whether additional task-specific anatomical knowledge is needed from the query, task specification, and compact policy context.
If needed, Anatomy-Aware Knowledge Retrieval is invoked; otherwise, the existing visual context is reused.
The controller then synthesizes an executable instruction for the selected action.
For generation, it uses the query and, when available, a selected visual support; for editing, it reasons over the retained best prediction, diagnosed errors, and relevant interaction trajectory to formulate a targeted correction.
The image model receives only the synthesized instruction and action-specific visual inputs.
Each new candidate is evaluated by Comparative Quality Critique, whose preference and residual-error diagnosis are incorporated into the structured state.
The retained best prediction is updated as
\begin{equation}
B_{t+1} =
\begin{cases}
\hat{M}_t, & \hat{M}_t \succ B_t,\\
B_t, & \text{otherwise},
\end{cases}
\end{equation}
where $\succ$ denotes the preference from comparative critique.
Without a previous best, the tournament winner initializes $B_t$.
This best prevents weaker generations or unsuccessful edits from overwriting stronger predictions.
Stage-specific prompts are provided in Sec. \ref{sec:prompt_details}.

\textbf{Adaptive Action Selection.}
Based on the current visual diagnosis, prior policy context, and interaction trajectory, the controller operates over an action space
$\mathcal{A}=\{\textsc{Generate},\textsc{Edit},\textsc{Accept}\}$.
\textsc{Generate} explores a new prediction when the retained result remains globally unreliable;
\textsc{Edit} refines the current best when it provides a reliable basis for local correction;
and \textsc{Accept} terminates inference once the result is satisfactory.
Otherwise, the updated state is carried to the next iteration, where knowledge needs and refinement strategy are reassessed.
Inference also stops when the iteration budget is exhausted, returning the best retained prediction.


\section{Experiments}
\label{sec:experiments}
In this section, we evaluate SegBanana on eight medical segmentation datasets and compare it with a broad range of baselines under a training-free setting. 
The datasets span diverse imaging modalities and segmentation targets:
TNBC \citep{naylor2018segmentation} for nuclei in histopathology,
RAVIR \citep{hatamizadeh2022ravir} for retinal vessels in fundus imaging,
Drishti-GS for optic disc and cup in fundus imaging,
ISIC for skin lesions in dermoscopy,
BUS-UCLM \citep{vallez2025bus} for breast lesions in ultrasound,
Kvasir \citep{jha2019kvasir} for gastrointestinal polyps in endoscopy,
ACDC for cardiac structures in MRI,
and BraTS \citep{menze2014multimodal,bakas2017advancing,baid2021rsna}
for brain tumors in multimodal MRI, with details in Sec. \ref{sec:dataset_details}.
We follow standard splits and report mean Dice (mDice) \citep{dice1945measures} as the primary metric. We use Nano Banana 2 for mask generation and editing, Gemini-3.5 as the multimodal controller, and DINOv3 ViT-L/16 with MMR ($\lambda=0.8$) for support retrieval. Each Generate action produces two candidates, each Edit action refines the current best prediction, and inference runs for up to three iterations with a compact structured state.

We compare against both medical-specific and generalist methods. Medical-specific baselines include UniverSeg \citep{butoi2023universeg}, BiomedParse \citep{biomedparse}, MedSAM3 \citep{liu2025medsam3}, IBISAgent \citep{jiang2026ibisagent}, and MedSAMAgent \citep{liu2026medsam}. Generalist baselines include open-vocabulary and in-context methods: Talk2DINO \citep{barsellotti2025talking}, SED \citep{xie2024sed}, MaskCLIP++ \citep{zeng2024maskclip++}, SAM3 \citep{carion2026sam}, Painter \citep{wang2023images}, SegGPT \citep{wang2023seggpt}, Matcher \citep{liu2023matcher}, PerSAM \citep{zhang2305personalize}, GF-SAM \citep{zhang2024bridge}, and INSID3 \citep{cuttano2026insid3}. We exclude SAM/SAM2-based methods requiring point or box prompts, as they introduce extra supervision unavailable in our setting.

\begin{table*}[t]
\centering
\caption{
Comparison of SegBanana against medical-specific and generalist baselines across eight medical segmentation datasets in terms of mDice (\%).
\zeromark/\fewmark denote zero-/few-shot settings.  \textcolor{gray!55}{Gray} denotes medical-specific results lower than SegBanana. \textbf{Bold}/\underline{underline} indicate the best/second-best results within each method category.
}
\label{tab:broad_generalist_comparison}
\resizebox{\textwidth}{!}{
\begin{tabular}{l|*{8}{c}|c}
\toprule
\textbf{Method}
& \textbf{TNBC}
& \textbf{RAVIR}
& \textbf{Drishti-GS}
& \textbf{ISIC}
& \textbf{BUS-UCLM}
& \textbf{Kvasir}
& \textbf{ACDC}
& \textbf{BraTS}
& \textbf{Avg}\\
\midrule

\multicolumn{10}{l}{\textit{Medical-Specific Methods}}\\
\midrule

UniverSeg\fewmark
& \textcolor{gray!55}{\underline{24.48}}
& \textcolor{gray!55}{\textbf{28.45}}
& \textcolor{gray!55}{40.68}
& \textcolor{gray!55}{39.64}
& \textcolor{gray!55}{13.38}
& \textcolor{gray!55}{19.91}
& \textcolor{gray!55}{21.48}
& \textcolor{gray!55}{15.48}
& \textcolor{gray!55}{25.44}\\

BiomedParse\zeromark
& \textcolor{gray!55}{11.81}
& \textcolor{gray!55}{\underline{5.67}}
& \textcolor{gray!55}{\textbf{78.19}}
& \underline{87.59}
& \textcolor{gray!55}{\textbf{63.81}}
& \textcolor{gray!55}{\underline{88.95}}
& \textbf{89.03}
& \textbf{75.11}
& \textcolor{gray!55}{\textbf{62.52}}\\

MedSAM3\zeromark
& \textcolor{gray!55}{\textbf{68.21}}
& \textcolor{gray!55}{4.21}
& \textcolor{gray!55}{34.13}
& \textbf{89.80}
& \textcolor{gray!55}{\underline{57.68}}
& \textbf{91.17}
& \textcolor{gray!55}{16.74}
& \textcolor{gray!55}{55.42}
& \textcolor{gray!55}{\underline{52.17}}\\

IBISAgent\zeromark
& \textcolor{gray!55}{1.00}
& \textcolor{gray!55}{1.65}
& \textcolor{gray!55}{4.14}
& 87.47
& \textcolor{gray!55}{57.06}
& \textcolor{gray!55}{78.86}
& \textcolor{gray!55}{14.87}
& \textcolor{gray!55}{\underline{60.14}}
& \textcolor{gray!55}{38.15}\\

MedSAMAgent\zeromark
& \textcolor{gray!55}{5.78}
& \textcolor{gray!55}{0.30}
& \textcolor{gray!55}{\underline{66.41}}
& 83.47
& \textcolor{gray!55}{49.16}
& \textcolor{gray!55}{83.92}
& \underline{80.50}
& \textcolor{gray!55}{21.38}
& \textcolor{gray!55}{48.87}\\

\midrule
\multicolumn{10}{l}{\textit{Generalist Methods}}\\
\midrule

Talk2DINO\zeromark
& 21.73
& 22.23
& 2.16
& 26.69
& 12.14
& 23.35
& 4.74
& 6.74
& 14.97\\

SED\zeromark
& 22.05
& 22.71
& 1.76
& 30.59
& 21.45
& 29.52
& 3.50
& 14.78
& 18.30\\

MaskCLIP++\zeromark
& 23.25
& 22.21
& 1.50
& 28.61
& 12.57
& 23.15
& 1.53
& 5.09
& 14.74\\

SAM3\zeromark
& 0.00
& 0.00
& 1.79
& 22.98
& 5.85
& 0.00
& \underline{56.65}
& 0.00
& 10.91\\

Painter\fewmark
& 22.94
& 6.55
& 1.72
& 27.65
& 8.05
& 27.60
& 1.36
& 15.55
& 13.93\\

SegGPT\fewmark
& \underline{53.33}
& \underline{64.19}
& \underline{76.82}
& 51.81
& 32.08
& 65.57
& 45.20
& \underline{29.08}
& \underline{52.26}\\

Matcher\fewmark
& 25.17
& 21.24
& 35.61
& 55.09
& 38.38
& 53.67
& 20.48
&{27.94}
& 34.70\\

PerSAM\fewmark
& 5.25
& 19.41
& 33.37
& 28.53
& 32.46
& 44.99
& 18.19
& 14.68
& 24.61\\

GF-SAM\fewmark
& 48.98
& 37.91
& 24.48
& 63.68
& \underline{42.07}
& 53.00
& 13.16
& 17.20
& 37.56\\

INSID3\fewmark
& 31.16
& 33.63
& 6.25
& \underline{63.87}
& 23.72
& \underline{74.49}
& 4.78
& 21.72
& 32.45\\

\midrule
\rowcolor{gray!10}
\textbf{SegBanana}\zeromark\fewmark
& \textbf{74.75}
& \textbf{80.93}
& \textbf{87.09}
& \textbf{82.06}
& \textbf{78.07}
& \textbf{89.53}
& \textbf{61.68}
& \textbf{65.52}
& \textbf{77.45}\\

\bottomrule
\end{tabular}
}
\end{table*}









\subsection{Main Results}
\label{sec:main_results}

\textbf{SegBanana achieves strong and balanced performance across diverse medical segmentation tasks.}
As shown in Tab.~\ref{tab:broad_generalist_comparison}, SegBanana achieves the highest average mDice of $77.45\%$, outperforming representative medical-specific and generalist baselines, BiomedParse and SegGPT, by $14.93$ and $25.19$ points, respectively.
It performs best on four datasets, including TNBC, RAVIR, Drishti-GS, and BUS-UCLM, and remains competitive with medical-specific methods on the other four.
Notably, several medical-specific methods outperforming SegBanana on these
datasets have dataset-level exposure to the corresponding benchmarks during
training (Sec.~\ref{sec:data_exposure}), underscoring the competitiveness of
SegBanana in a training-free setting.
Across method families, medical-specific models perform strongly within their training coverage but degrade beyond it, generalist methods may lack specialized anatomical knowledge and suffer from foreground--background ambiguity and coarse patch-level matching, limiting fine-structure segmentation.
In contrast, SegBanana remains consistently strong across all datasets.

\begin{figure}[tb]
  \centering
  \includegraphics[width=1\textwidth]{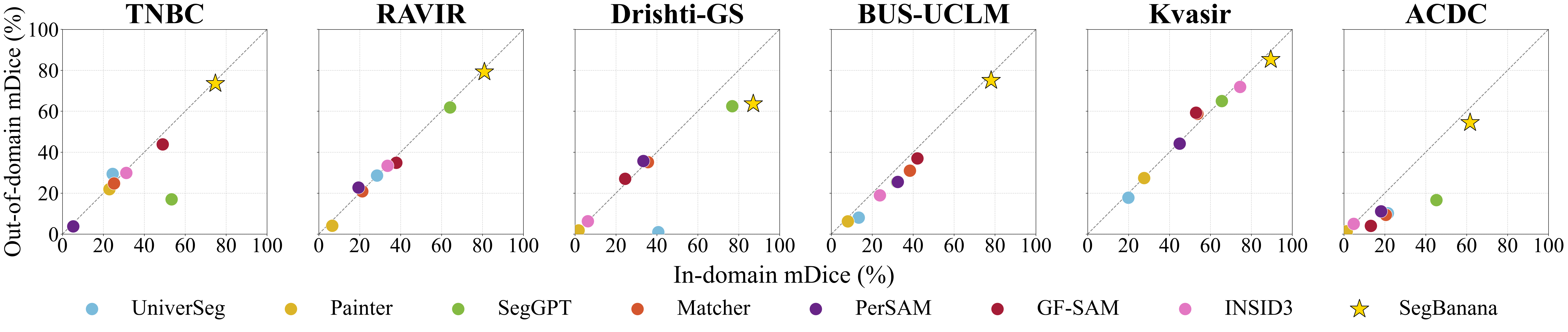}
  \vspace{-2em}
  \caption{In-domain vs. out-of-domain mDice (\%) across datasets. Points closer to the diagonal indicate greater robustness to support-domain shift.
  }
  \label{fig:ood}
\end{figure}
\textbf{SegBanana remains robust when in-domain supports are unavailable.}
New clinical scenarios may lack matched annotated supports, while related external datasets with similar targets remain available. We therefore evaluate six query datasets using suitable public out-of-domain supports and compare representative few-shot methods under the same setting, with dataset pairings provided in Sec. \ref{sec:cross_domain_support}. As shown in Fig.~\ref{fig:ood}, points near the diagonal indicate robustness to support-domain shift. SegBanana remains close to the diagonal on most datasets, whereas several baselines degrade substantially. For example, SegGPT drops from $53\%$ to $17\%$ on TNBC and from $45\%$ to $16\%$ on ACDC. 
These results highlight the potential of SegBanana to reduce the annotation burden for deploying medical segmentation in new clinical practice.

\subsection{Ablation}
\begin{table*}[t]
\centering
\caption{
Ablation of framework components and multimodal controller backbones across eight datasets. Results are reported in mDice (\%).
}
\label{tab:component_ablation}
\setlength{\tabcolsep}{3.5pt}
\resizebox{\textwidth}{!}{
\begin{tabular}{lc|cccccccc|c}
\toprule
\textbf{Framework}
& \textbf{Controller}
& \textbf{TNBC}
& \textbf{RAVIR}
& \textbf{Drishti-GS}
& \textbf{ISIC}
& \textbf{BUS-UCLM}
& \textbf{Kvasir}
& \textbf{ACDC}
& \textbf{BraTS}
& \textbf{Avg.} \\
\midrule

SegBanana
& Gemini-3.5
& 74.75
& 80.93
& 87.09
& 82.06
& 78.07
& 89.53
& 61.68
& 65.52
& 77.45 \\

\quad w/o Retrieval
& Gemini-3.5
& 76.26
& 81.22
& 60.55
& 82.78
& 77.92
& 88.19
& 51.56
& 61.67
& 72.52 \\

\quad w/o Critic
& Gemini-3.5
& 72.08
& 78.05
& 63.88
& 81.16
& 75.07
& 86.02
& 56.68
& 65.66
& 72.33 \\

\quad w/o State Context
& Gemini-3.5
& 73.86
& 77.30
& 65.07
& 81.85
& 76.18
& 87.74
& 56.36
& 63.43
& 72.72 \\

\midrule

SegBanana
& Qwen-3.8
& 70.06
& 41.94
& 83.28
& 79.03
& 69.37
& 82.31
& 51.31
& 54.32
& 66.45 \\

SegBanana
& GPT-5.6
& 62.38
& 54.36
& 88.15
& 82.78
& 77.54
& 85.39
& 55.39
& 61.88
& 70.98 \\

\midrule

Nano Banana 2 & N/A
& 67.04
& 34.44
& 43.73
& 74.68
& 69.30
& 85.06
& 36.89
& 50.38
& 57.69 \\
\bottomrule
\end{tabular}
}
\vspace{-0.5em}
\end{table*}

\textbf{SegBanana consistently benefits from its key components and remains robust to the choice of multimodal controller.}
We assess the contribution of each framework component and the robustness of SegBanana to multimodal controllers. As shown in Tab.~\ref{tab:component_ablation}, \emph{w/o retrieval}, \emph{w/o critic}, and \emph{w/o state context} reduce average mDice from $77.45\%$ to $72.52\%$, $72.33\%$, and $72.72\%$, respectively. Here, \emph{w/o retrieval} forces all cases to skip support retrieval; \emph{w/o critic} integrates comparative evaluation into the controller rather than using a separate critic tool; and \emph{w/o state context} replaces the compact structured state with the full interaction context. Retrieval is particularly important on knowledge-intensive datasets such as Drishti-GS and ACDC, while the critic and state ablations demonstrate the benefits of dedicated comparative evaluation and compact history-aware reasoning. SegBanana also remains effective with Qwen and GPT controllers despite variation across backbones. Finally, direct generation with Nano Banana 2 achieves only $57.69\%$ average mDice, highlighting the substantial gain brought by agentic inference over the frozen generator alone.
\subsection{Discussion}
\begin{figure}[tb]
  \centering
\includegraphics[width=1\textwidth]{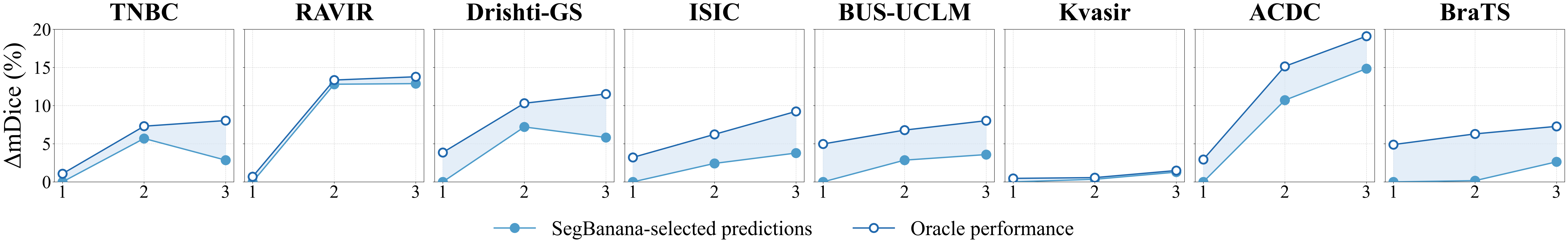}
\vspace{-2em}
  \caption{Iterative segmentation performance across inference steps, with the $\Delta$mDice of critique-selected predictions and the oracle performance among all masks generated up to each iteration.
  }
  \vspace{-1em}
  \label{fig:ablation}
\end{figure}

\textbf{Closed-loop inference progressively improves segmentation performance, but Comparative Quality Critique remains limited in consistently identifying the best generated candidate.}
Fig.~\ref{fig:ablation} shows that closed-loop inference steadily expands the pool of strong predictions, as reflected by the consistently improving oracle performance. While SegBanana allows up to three iterations, its actual inference depth and action usage are adaptively determined by the evolving prediction state, with detailed efficiency and stopping statistics provided in Sec.~\ref{sec:efficiency}. However, the prediction selected by Comparative Quality Critique remains below the oracle across iterations, indicating that better candidates are often generated but not reliably identified. This gap highlights candidate selection as a key remaining bottleneck and suggests that improving training-free comparative evaluation could further unlock the gains already available from iterative generation and refinement.

\begin{figure}[tb]
  \centering
\includegraphics[width=1\textwidth]{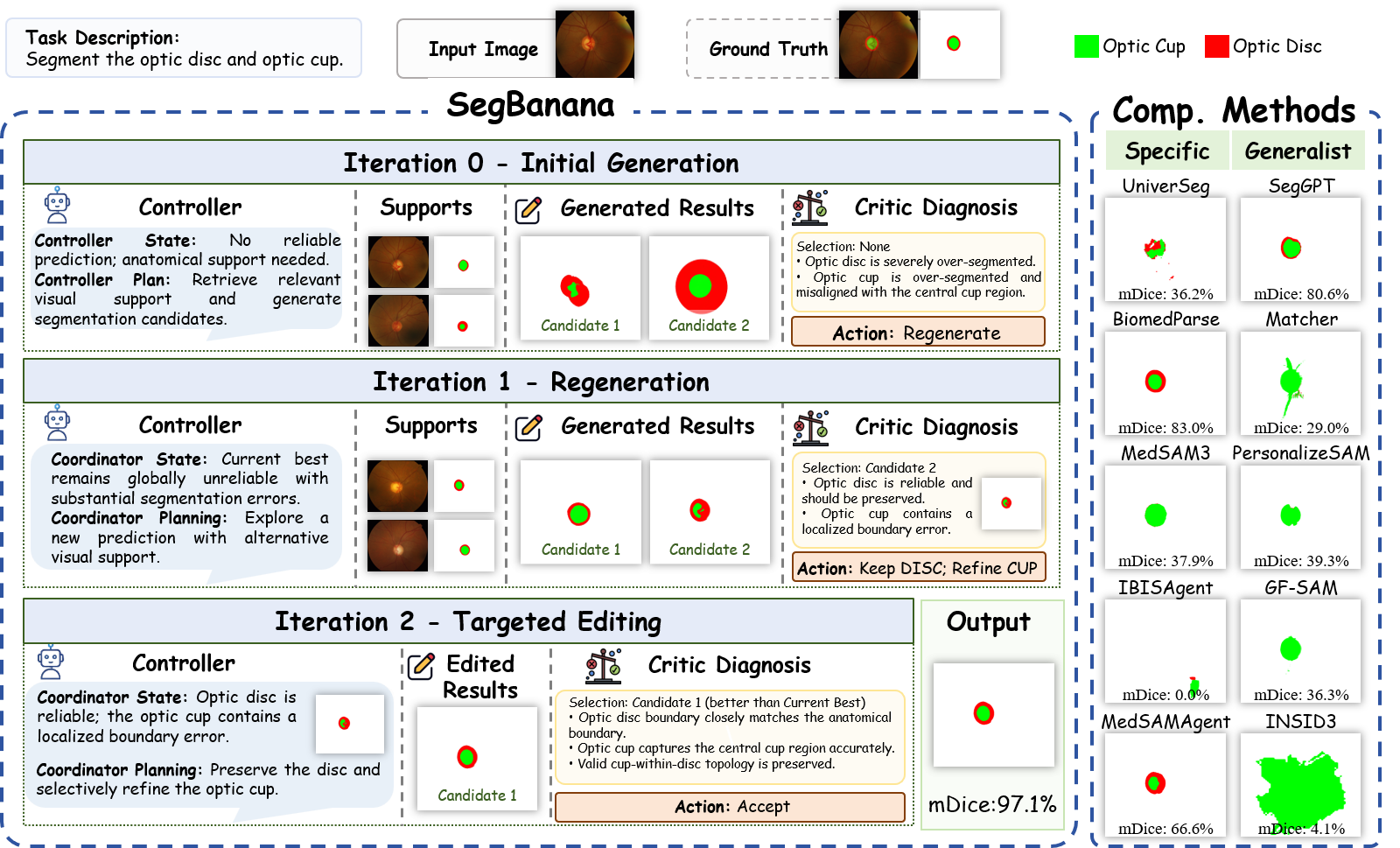}
\vspace{-2em}
  \caption{Case study of SegBanana, illustrating its iterative segmentation process and comparison with the top five methods from medical-specific and generalist baselines.
}
\vspace{-1em}
  \label{fig:case}
\end{figure}

\textbf{Qualitative Study.}
Fig.~\ref{fig:case} illustrates how SegBanana benefits from adaptive, state-aware inference. When the initial prediction suffers severe over-segmentation, the critic triggers regeneration rather than local editing. Once a stronger candidate is obtained, the controller preserves the reliable optic disc and selectively refines the remaining cup error, yielding a final mDice of $97.1\%$. This trajectory illustrates how the controller adapts its action policy as the state evolves, switching from global exploration to local refinement once a reliable hypothesis emerges. Additional case studies across diverse datasets are provided in Sec.~\ref{sec:additional_cases}.

\section{Conclusion}
In this paper, we present SegBanana, an agentic visual generation framework that enables a frozen UMM to perform medical image segmentation without task-specific post-training. SegBanana harnesses intrinsic generation and editing capabilities through retrieval, refinement, and verification at inference time. Across diverse datasets, it outperforms medical-specific and generalist baselines and remains robust without in-domain supports, demonstrating effective training-free transfer. A key \textbf{limitation} is the imperfect reliability of comparative quality critique, which prevents SegBanana from fully exploiting the UMM's segmentation potential. \textbf{Future work} will pursue stronger verification and more autonomous agentic control, including a dedicated policy model for reasoning over intermediate outcomes and orchestrating visual actions.


\section*{AI Use Statement}
For manuscript preparation, we use OpenAI GPT-family models only for language
polishing and grammar correction. Generative and multimodal models used as
experimental components of SegBanana are explicitly described in the paper.
AI tools were not used for literature review or idea formation.

\section*{Ethics Statement}

This work uses only publicly available medical image segmentation datasets, and no new patient data were collected, no human subjects were recruited, and no patient interaction occurred. We do not attempt to identify or re-identify individuals from the medical images. SegBanana is intended solely for research purposes and is not a medical device. Although the framework demonstrates strong transfer across diverse datasets, segmentation errors may still arise under challenging anatomy, distribution shifts, or imperfect candidate evaluation. We therefore report methodological limitations and do not claim clinical readiness.

\section*{Reproducibility Statement}
To facilitate reproducibility, we provide detailed implementation settings in Sec. \ref{sec:experiments} and the supplementary material, including model configurations, stage-specific prompts and input--output formats, support retrieval, structured state construction, agentic rollouts, and hyperparameters. We specify the UMM and multimodal controller used in each experiment, the DINOv3-based retrieval and MMR settings, the number of candidates generated per action, and the maximum number of inference iterations. Upon publication, we will release the code, prompt templates, configuration files, dataset splits, support-pool construction, and preprocessing pipeline. Because SegBanana relies on externally hosted UMM APIs for generation, editing, and multimodal reasoning, exact outputs may vary slightly due to stochastic inference or future API/model updates. To mitigate this source of variation, we will document the exact model versions and inference settings used for all reported results.

\bibliography{iclr2027_conference}
\bibliographystyle{iclr2027_conference}

\newpage
\appendix

\section{Limitations and Future Work}
SegBanana currently has two main limitations. First, the framework operates on 2D images and cannot directly model volumetric 3D context, requiring 3D scans to be processed slice by slice. Second, although Comparative Quality Critique enables training-free candidate selection and refinement, its reliability remains imperfect, preventing the framework from fully exploiting the segmentation capability of pretrained UMMs. Future work could incorporate targeted supervision to strengthen both the controller and the UMM on medical images, while preserving their pretrained visual understanding, reasoning, and generation abilities. Such adaptation may further improve medical segmentation performance without sacrificing generality, moving toward a more flexible and broadly transferable segmentation paradigm.

\section{Appendix}
\subsection{Dataset Details and Coverage}
\label{sec:dataset_details}
\begin{table*}[t]
\centering
\caption{
Characteristics of the evaluation datasets in terms of imaging modality, segmentation target, spatial structure, and primary segmentation challenge.
}
\label{tab:dataset_coverage}
\resizebox{\textwidth}{!}{
\begin{tabular}{lccll}
\toprule
\textbf{Dataset}
& \textbf{Modality}
& \textbf{Target}
& \textbf{Structure}
& \textbf{Primary Challenge} \\
\midrule
ACDC
& MRI
& LV / RV / MYO
& Adjacent
& Anatomical coupling \\
Drishti-GS
& Fundus
& Optic disc / cup
& Nested
& Subtle boundary discrimination \\
ISIC
& Dermoscopy
& Skin lesions
& Irregular
& Appearance variation \\
TNBC
& Histopathology
& Cell nuclei
& Irregular, Adjacent
& Dense clustering \& appearance variation \\
RAVIR
& Fundus
& Retinal vessels
& Irregular
& Subtle boundaries \& topology preservation \\
BUS-UCLM
& Ultrasound
& Breast lesions
& Irregular
& Appearance variation \& weak boundaries \\
Kvasir
& Endoscopy
& Colorectal polyp
& Irregular
& Appearance variation \& specular reflection \\
BraTS
& MRI
& Brain tumor
& Nested
& Anatomical coupling \& appearance variation \\
\bottomrule
\end{tabular}
}
\end{table*}
We evaluate SegBanana on eight medical image segmentation datasets covering diverse imaging modalities, anatomical targets, spatial scales, and structural characteristics.
As summarized in Tab.~\ref{tab:dataset_coverage}, these datasets span MRI, fundus imaging, dermoscopy, histopathology, ultrasound, and endoscopy, with targets ranging from small cellular and vascular structures to large lesions and multi-region anatomical structures.
They also exhibit substantially different spatial organizations, including adjacent, nested, and irregular structures, resulting in distinct segmentation challenges such as subtle boundaries, anatomical coupling, dense object clustering, topology preservation, and large appearance variation.
Such diversity allows us to evaluate not only cross-modality generalization, but also the ability to transfer across segmentation tasks with different requirements for appearance recognition, boundary discrimination, and anatomical reasoning.

\subsection{Supervision Profile of Compared Methods}
\label{sec:data_exposure}
\begin{table*}[t]
\centering
\caption{
Reported medical data exposure of the compared medical-specific methods.
Datasets used during any model-development stage, including pretraining,
fine-tuning/SFT, and RL, are counted as exposure.
Gray rows denote modalities included in our evaluation benchmark, while
\textcolor{overlapred}{red} dataset names indicate benchmark-level overlap
with our evaluation datasets.
$\ddagger$ indicates that exposure to the modality is explicitly reported,
but the complete dataset-level training list is not publicly disclosed.
}
\label{tab:data_exposure}

\scriptsize
\setlength{\tabcolsep}{2.8pt}
\renewcommand{\arraystretch}{1.22}

\begin{tabularx}{\textwidth}{
    l
    >{\raggedright\arraybackslash}X
    >{\raggedright\arraybackslash}X
    >{\raggedright\arraybackslash}X
    >{\raggedright\arraybackslash}X
    >{\raggedright\arraybackslash}X
}
\toprule
\textbf{Modality}
& \textbf{UniverSeg}
& \textbf{BiomedParse}
& \textbf{MedSAM3}
& \textbf{IBISAgent}
& \textbf{MedSAM-Agent} \\
\midrule

\textbf{MRI}
&
\overlap{BraTS}, BrainDevelopment, ISLES, LGG, MCIC,
OASIS, PPMI, WMH, I2CVB, NCI-ISBI, PROMISE12,
FeTA, SpineWeb, \overlap{ACDC}, CDEMRIS
&
\overlap{BraTS2023}, PROMISE12, LGG,
\overlap{ACDC}, M\&Ms
&
MRI corpus$^{\ddagger}$
&
\overlap{ACDC}, LGG, M\&Ms,
MSD Brain Tumor, MSD Heart,
MSD Hippocampus, MSD Prostate
&
\overlap{ACDC}, LGG, AMOS-MRI
\\
\midrule

\textbf{CT}
&
AbdomenCT-1K, BTCV, KiTS,
LiTS, LUNA, SegTHOR, WORD
&
AbdomenCT-1K, BTCV, KiTS23,
TotalSegmentator, COVID-19 CT,
LIDC-IDRI, FUMPE
&
CT corpus$^{\ddagger}$
&
COVID-19 CT, KiTS23, LIDC-IDRI,
MSD Liver, MSD Spleen, MSD Pancreas,
MSD Colon, MSD Lung, MSD Hepatic Vessel
&
FLARE22, KiTS, LIDC-IDRI,
BTCV, AMOS-CT, WORD
\\
\midrule

\textbf{MRI \& CT}
&
AMOS, CHAOS, MSD
&
AMOS22, MSD
&
--
&
AMOS22, MSD task groups
&
--
\\
\midrule

\textbf{Fundus / Retinal}
&
DRIVE, STARE, IDRID, e-Ophtha
&
DRIVE, REFUGE, G1020
&
RIM-ONE
&
G1020, REFUGE
&
REFUGE
\\
\midrule

\textbf{X-ray}
&
CheXplanation, PanDental
&
COVID-QU-Ex, QaTa-COV19,
SIIM-ACR Pneumothorax,
Chest Xray Masks and Labels,
CDD-CESM, Radiography
&
X-ray corpus$^{\ddagger}$
&
COVID-QU-Ex, QaTa-COV19,
Radiography, CDD-CESM,
SIIM-ACR Pneumothorax,
CXR Masks and Labels
&
CXRMask, Radiography series,
CDD-CESM
\\
\midrule

\textbf{Ultrasound}
&
CAMUS, HMC-QU, BUS, TUCC
&
CAMUS, BUSI,
US Simulation \& Segmentation,
BreastUS, LiverUS, FH-PS-AOP
&
BUSI; broader US corpus$^{\ddagger}$
&
CAMUS, BreastUS,
LiverUS, FH-PS-AOP
&
BreastUS, LiverUS, FH-PS-AOP
\\
\midrule

\textbf{Histopathology / Microscopy}
&
BBBC003, WBC, CoNSeP
&
PanNuke, GlaS
&
Histopathology / microscopy corpus$^{\ddagger}$
&
GlaS, PanNuke
&
--
\\
\midrule

\textbf{Dermoscopy}
&
--
&
\overlap{ISIC 2018}, UWaterlooSkinCancer
&
\overlap{ISIC 2018}; broader corpus$^{\ddagger}$
&
\overlap{ISIC}, UWaterlooSkinCancer
&
--
\\
\midrule

\textbf{Endoscopy}
&
--
&
PolypGen, NeoPolyp
&
\overlap{Kvasir-SEG}; broader corpus$^{\ddagger}$
&
NeoPolyp, PolypGen
&
NeoPolyp, PolypGen
\\
\midrule

\textbf{OCT / OCTA}
&
OCTA500, ROSE
&
OCT-CME
&
OCT corpus$^{\ddagger}$
&
OCT-CME
&
--
\\

\bottomrule
\end{tabularx}

\vspace{1mm}
\begin{minipage}{\textwidth}
\footnotesize
\textbf{Our evaluation datasets.}
MRI: ACDC and BraTS;
Fundus: Drishti-GS and RAVIR;
Ultrasound: BUS-UCLM;
Histopathology: TNBC;
Dermoscopy: ISIC;
Endoscopy: Kvasir.
For MedSAM3, the public release reports broader multimodal medical training
than the individually named datasets; therefore, modality-level exposure is
reported where the complete dataset list is unavailable.
\end{minipage}

\end{table*}
Medical-specific baselines differ substantially in the amount and diversity of
medical data observed during model development.
To better contextualize their performance, we summarize the publicly reported
medical data exposure of each method in Tab.~\ref{tab:data_exposure}.
We consider a dataset as ``seen'' whenever it is explicitly used at any stage
of model development, including pretraining, supervised fine-tuning (SFT),
or reinforcement learning (RL), since exposure at any of these stages may
influence downstream segmentation behavior.
Dataset names highlighted in \overlap{red} indicate benchmark-level overlap with our evaluation datasets. This notation indicates exposure to the same benchmark
and does not necessarily imply identical training/test samples or splits.

Overall, the compared medical-specific methods exhibit markedly different
training exposure. Several methods have observed datasets from the same
modalities, anatomical regions, or even the same benchmarks used in our
evaluation. Such prior exposure provides useful context for interpreting
dataset-specific performance and out-of-domain generalization.
For methods whose complete foundation-model pretraining corpora are not
publicly disclosed, we report only explicitly documented medical datasets
or modality-level exposure and make no assumptions about undisclosed data.

\subsection{Out-of-Domain Support Construction}
\label{sec:cross_domain_support}
\begin{figure}[tb]
  \centering
\includegraphics[width=1\textwidth]{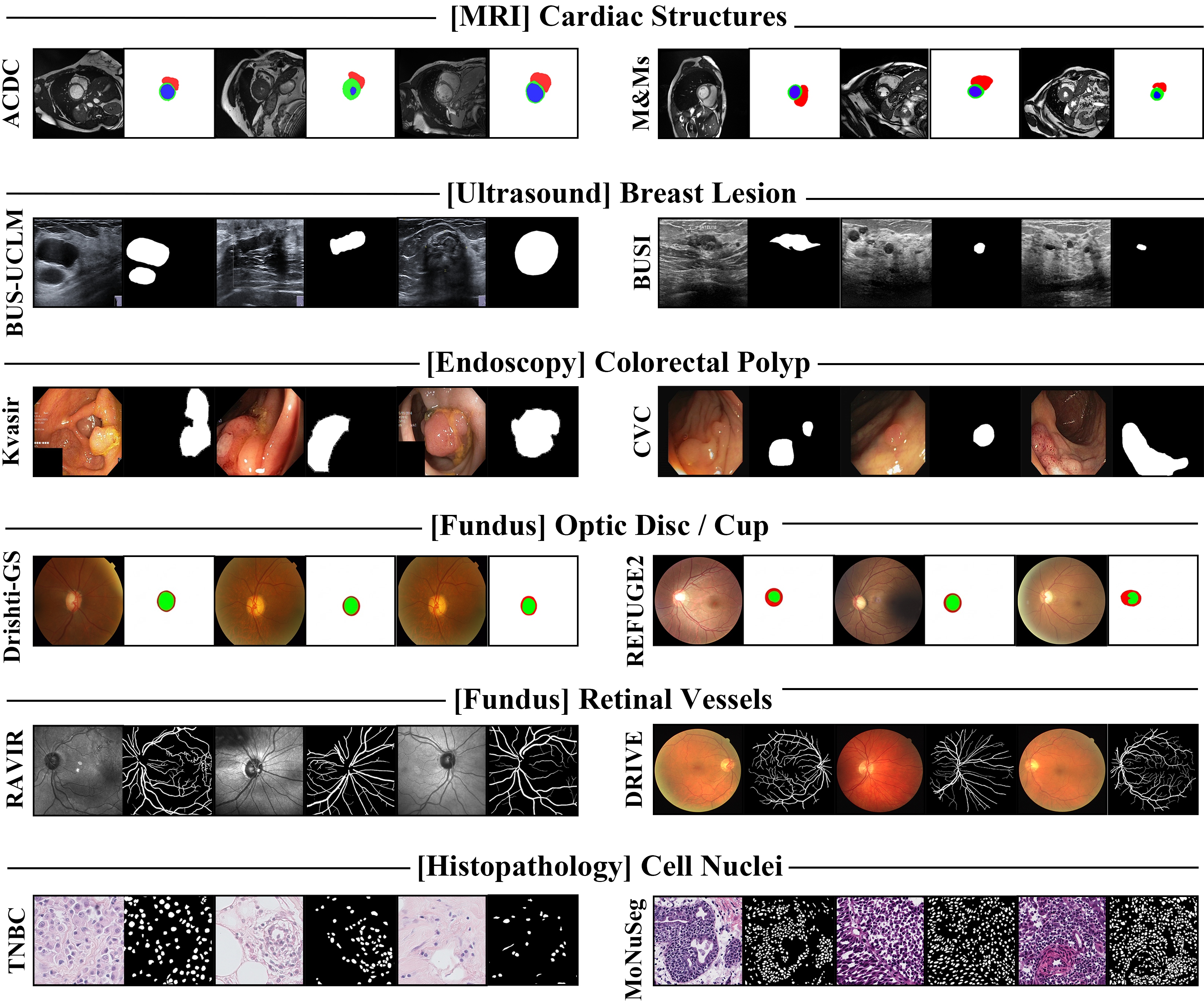}
  \caption{
Out-of-domain support construction across six medical segmentation tasks.
For each task, the query dataset (left) is paired with an external annotated dataset (right) sharing the same or closely corresponding segmentation target, while differing in acquisition conditions, appearance, population, or annotation characteristics.
Example image--mask pairs illustrate the resulting domain shift while preserving task-relevant anatomical semantics.
}
  \label{fig:cross_domain_support}
\end{figure}
To evaluate SegBanana when matched in-domain annotations are unavailable, we construct an out-of-domain support setting in which each query dataset is paired with an external annotated dataset containing the same or closely corresponding
segmentation target. Importantly, the external supports are drawn from a different dataset rather than from the query distribution, introducing shifts in acquisition conditions, appearance, population, and annotation characteristics while preserving the
task-relevant anatomical semantics. The dataset pairings are visualized in
Fig.~\ref{fig:cross_domain_support} to illustrate the corresponding domain gaps.

\subsection{Prompt and Inference Details}
\label{sec:prompt_details}

SegBanana decomposes inference into several stage-specific prompting modules, including knowledge-sufficiency assessment, generation and editing instruction synthesis, visual execution, comparative critique, and action selection. Generation and editing each contain two stages: the controller first synthesizes an executable instruction from the relevant interaction state, and the frozen UMM then executes this instruction using only the visual inputs required by the selected action. Different stages therefore receive different projections of the structured state rather than the complete interaction trajectory. Controller-side prompts use fixed JSON output formats for deterministic parsing, while generation and editing execution directly output RGB segmentation masks.

\paragraph{Task Description.}
Each dataset is associated with a concise segmentation instruction defining the target structures, as summarized in Tab.~\ref{tab:task_descriptions}. These task descriptions are further instantiated with dataset-specific label definitions and RGB conventions when constructing the corresponding prompts.

\begin{table}[t]
\centering
\caption{Dataset-specific task descriptions used in prompts.}
\label{tab:task_descriptions}
\small
\begin{tabular}{ll}
\toprule
\textbf{Dataset} & \textbf{Task Description} \\
\midrule
ACDC &
Segment the right ventricular cavity, myocardium, and left ventricular cavity. \\

BUS-UCLM &
Segment the breast lesion or mass. \\

Kvasir &
Segment the gastrointestinal polyp. \\

RAVIR &
Segment the retinal vessels. \\

BraTS &
Segment the brain tumor. \\

TNBC &
Segment the cell nuclei. \\

Drishti-GS &
Segment the optic disc and optic cup. \\

ISIC &
Segment the skin lesion. \\
\bottomrule
\end{tabular}
\end{table}

\begin{figure}[tb]
  \centering
  \includegraphics[width=1\textwidth]{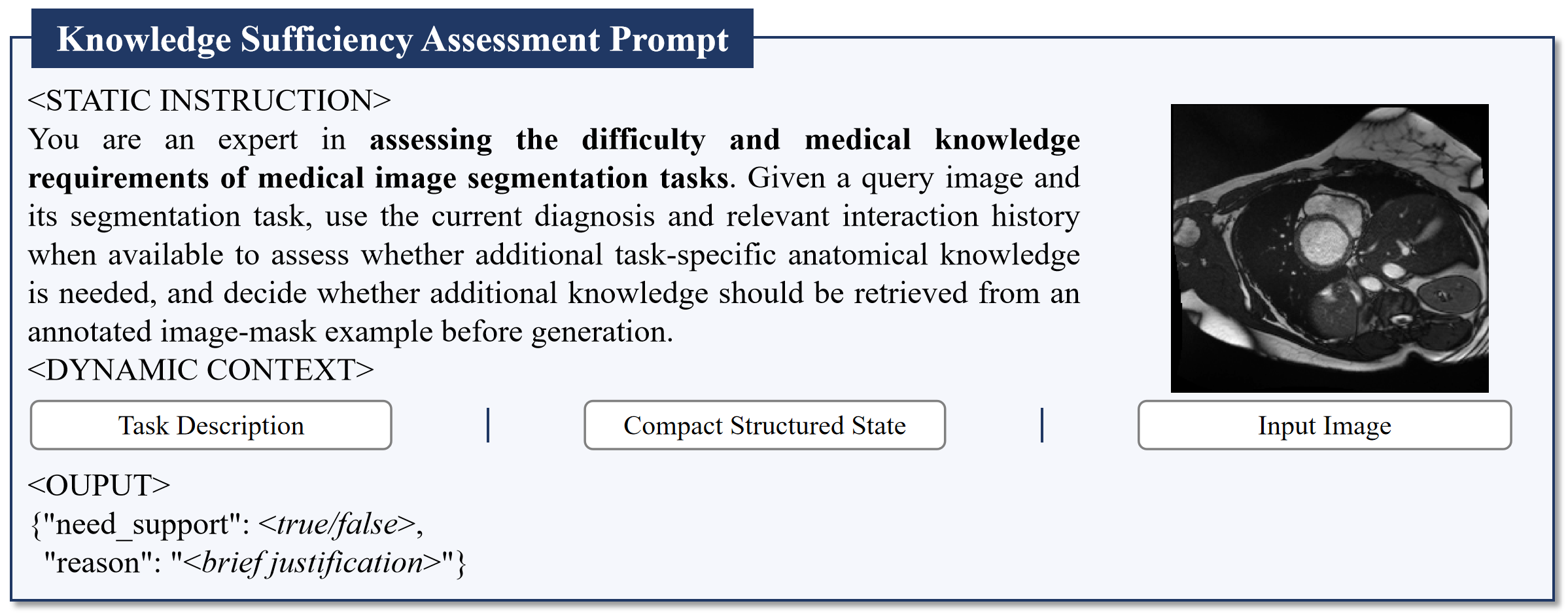}
  \caption{
  Knowledge Sufficiency Assessment prompt. The controller assesses whether additional task-specific anatomical knowledge is required from the task description, query image, and compact structured state, and returns a fixed JSON decision indicating whether support retrieval is needed.
  }
  \label{fig:p_retrieval}
\end{figure}
\paragraph{Knowledge Sufficiency Assessment.}
Before each generation step, the controller determines whether the available context contains sufficient task-specific anatomical knowledge. As illustrated in Fig.~\ref{fig:p_retrieval}, the prompt receives the task description, query image, and compact structured state, and outputs a structured decision indicating whether an additional annotated support example should be retrieved. The retriever itself is feature-based and does not use a language-model prompt.

\begin{figure}[tb]
  \centering
  \includegraphics[width=1\textwidth]{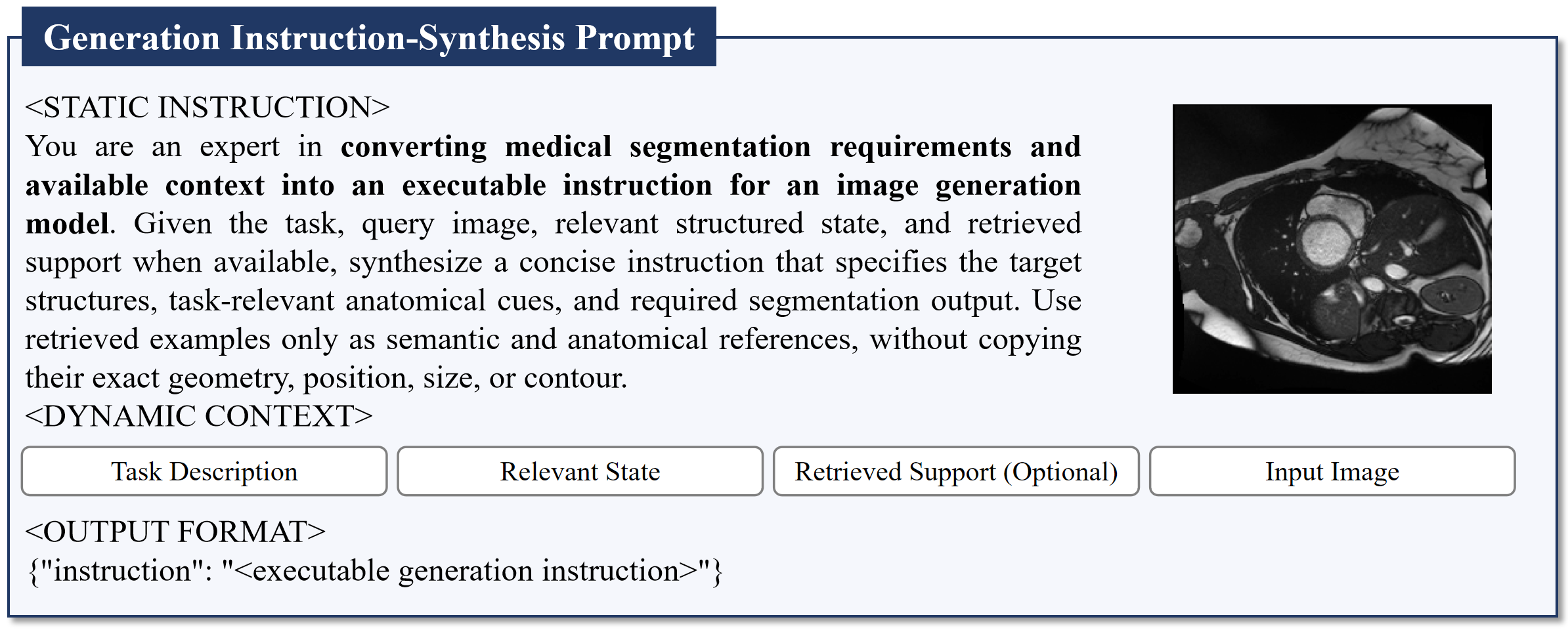}
  \caption{
  Generation Instruction-Synthesis prompt. The controller combines the task description, stage-relevant structured state, query image, and optional retrieved support to produce an executable generation instruction for the frozen UMM.
  }
  \label{fig:p_gen_ins}
\end{figure}
\paragraph{Generation Instruction Synthesis.}
When \textsc{Generate} is selected, the controller converts the task specification and relevant context into an executable instruction for the frozen UMM. As shown in Fig.~\ref{fig:p_gen_ins}, the prompt may additionally include a retrieved image--mask support pair. Retrieved examples are used only to provide semantic, anatomical, and appearance cues, rather than as geometric templates to be copied. The controller outputs the synthesized generation instruction in a fixed JSON format.

\begin{figure}[tb]
  \centering
  \includegraphics[width=1\textwidth]{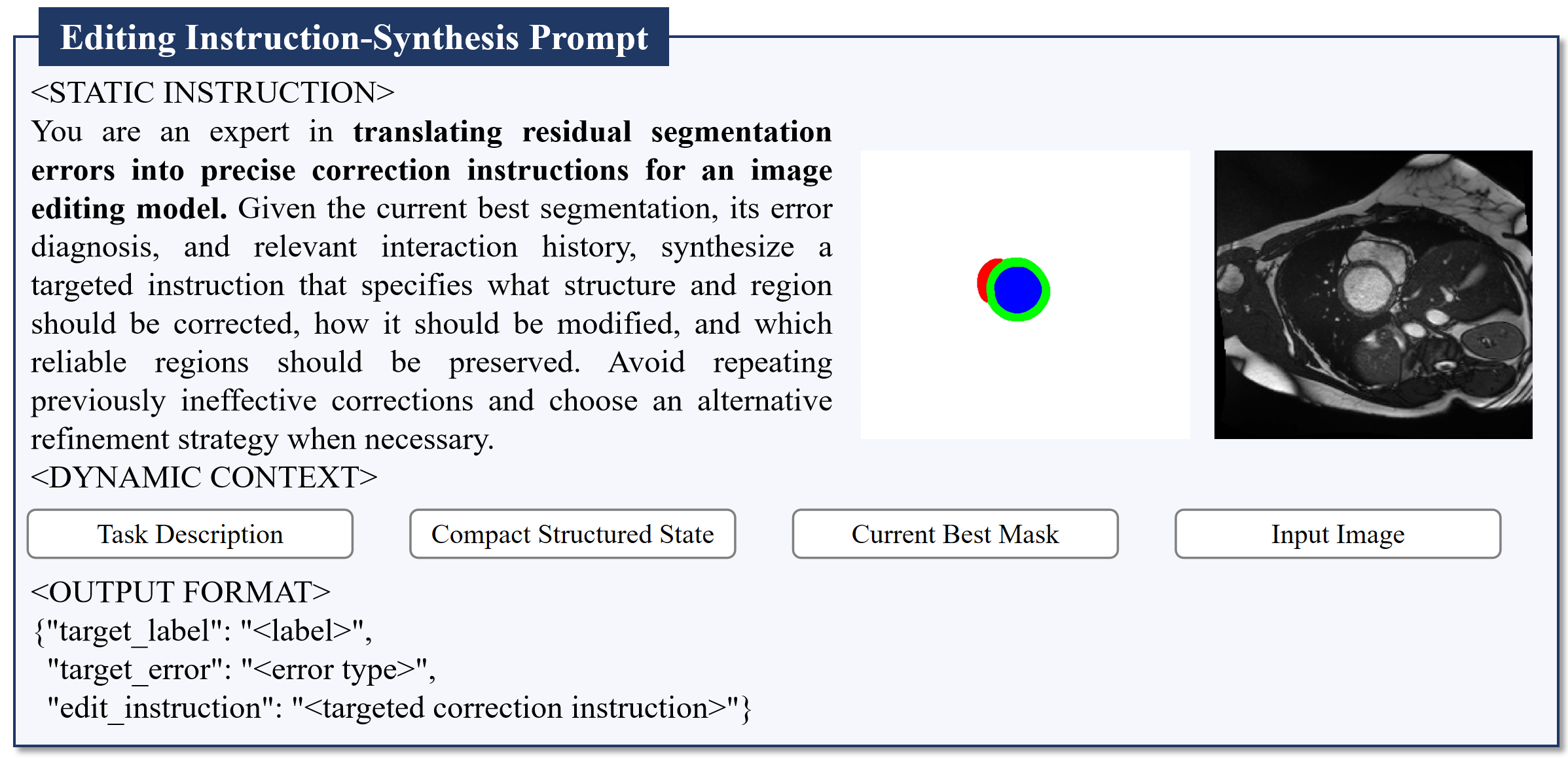}
  \caption{
  Editing Instruction-Synthesis prompt. The controller transforms the retained prediction, residual-error diagnosis, and relevant interaction trajectory into a targeted correction instruction specifying the structure, error, and desired local revision.
  }
  \label{fig:p_edit_ins}
\end{figure}

\paragraph{Editing Instruction Synthesis.}
When the retained prediction provides a reliable basis for local correction, the controller synthesizes a targeted editing instruction. As shown in Fig.~\ref{fig:p_edit_ins}, the controller reasons over the query image, retained best mask, residual-error diagnosis, and relevant editing history to determine what structure should be corrected, how it should be modified, and which reliable regions should be preserved. Previously ineffective corrections are also considered to avoid repeating failed refinement strategies.

\begin{figure}[tb]
  \centering
  \includegraphics[width=1\textwidth]{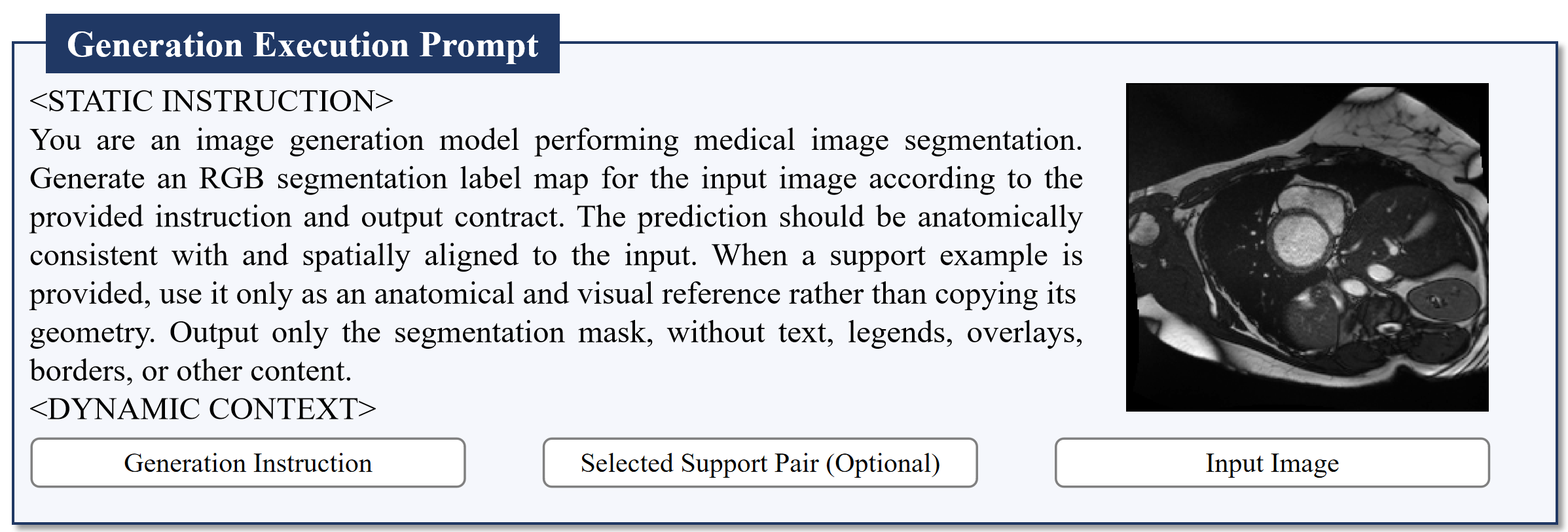}
  \caption{
  Generation Execution prompt. The frozen UMM receives the query image, synthesized generation instruction, optional selected support pair, and output contract, and directly generates an RGB segmentation mask.
  }
  \label{fig:p_gen}
\end{figure}
\paragraph{Generation Execution.}
The synthesized generation instruction is then executed by the frozen UMM. As illustrated in Fig.~\ref{fig:p_gen}, the image model receives the query image, the synthesized instruction, an optional selected support pair, and the dataset-specific output contract. It is instructed to produce only a spatially aligned RGB segmentation label map following the predefined label-color convention.

\begin{figure}[tb]
  \centering
  \includegraphics[width=1\textwidth]{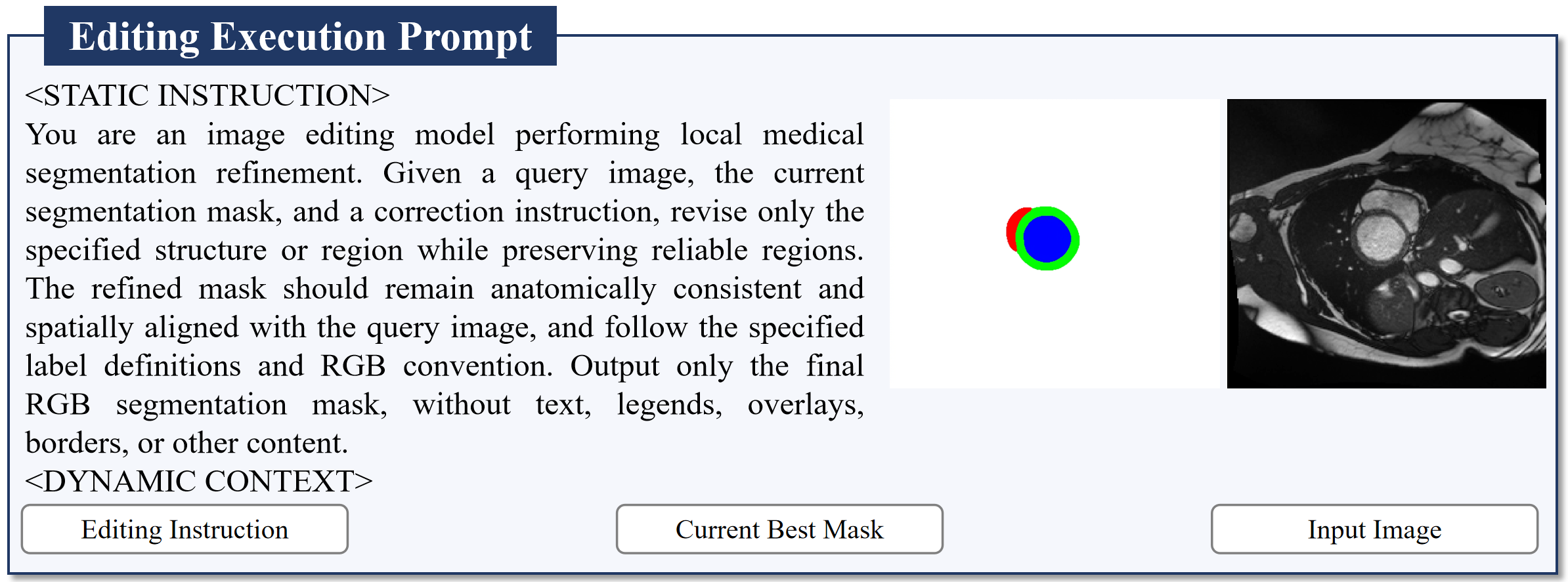}
  \caption{
  Editing Execution prompt. The frozen UMM locally revises the retained best mask using the query image and synthesized correction instruction while preserving reliable regions and following the output contract.
  }
  \label{fig:p_edit}
\end{figure}
\paragraph{Editing Execution.}
Editing uses the same frozen UMM but a different visual context. As shown in Fig.~\ref{fig:p_edit}, the image model receives only the query image, retained best mask, targeted editing instruction, and output contract. Structured interaction trajectory and retrieved supports are not directly exposed to the editor; their relevant information has already been incorporated into the synthesized correction instruction.

\begin{figure}[tb]
  \centering
  \includegraphics[width=1\textwidth]{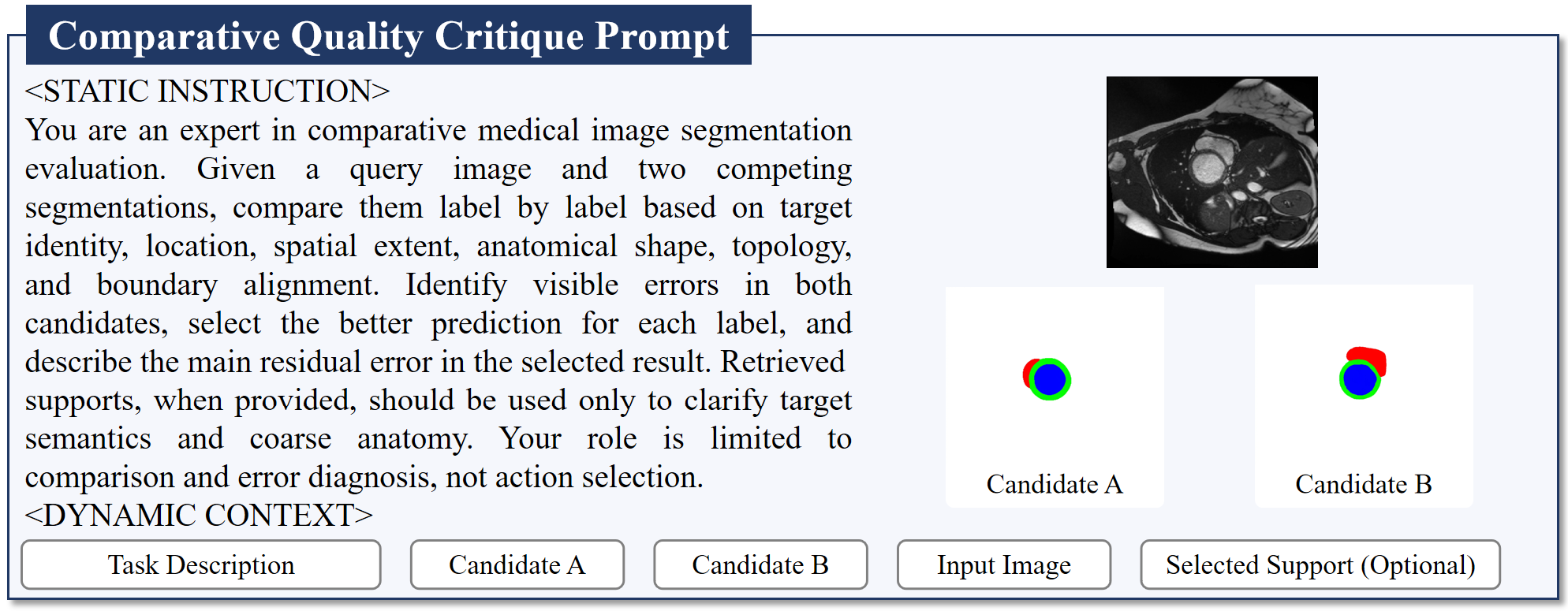}
  \caption{
  Comparative Quality Critique prompt. A frozen multimodal critic compares a baseline or current champion against a challenger, performs label-wise quality assessment and residual-error diagnosis, and returns the comparison in a fixed structured format.
  }
  \label{fig:p_critic}
\end{figure}
\paragraph{Comparative Quality Critique.}
As illustrated in Fig.~\ref{fig:p_critic}, the critic performs pairwise visual comparison rather than independent absolute scoring. Given the query image and two competing predictions, it compares segmentation quality label by label according to target identity, location, spatial extent, shape, topology, and boundary alignment. Optional retrieved supports may be provided only to clarify target semantics and coarse anatomy. The critic returns a structured preference together with label-wise error diagnosis, but does not determine the next inference action. Sequential pairwise comparisons are used to obtain the current tournament winner.

\begin{figure}[tb]
  \centering
  \includegraphics[width=1\textwidth]{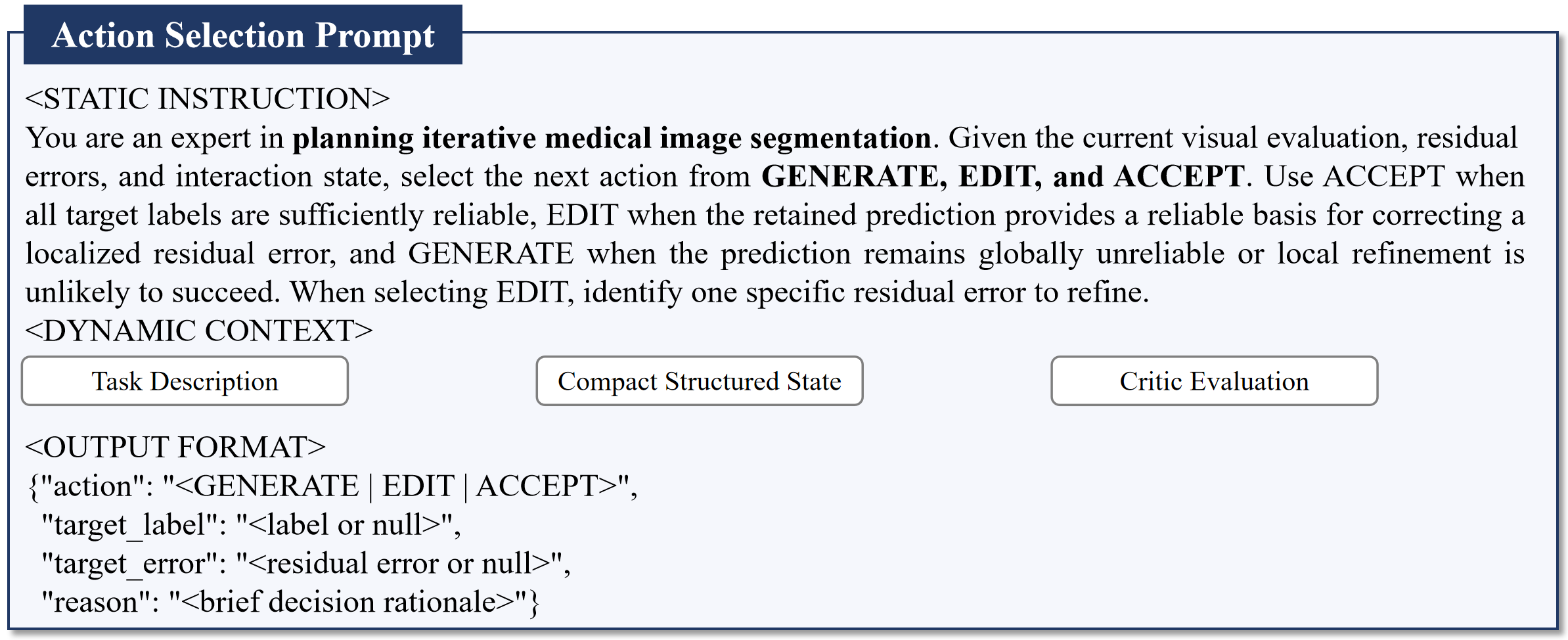}
  \caption{
  Action Selection prompt. Based on the compact structured state and comparative critique, the controller selects \textsc{Generate}, \textsc{Edit}, or \textsc{Accept}, and specifies the target residual error when editing is chosen.
  }
  \label{fig:p_action}
\end{figure}

\paragraph{Adaptive Action Selection.}
Finally, the controller determines the next inference action from the current visual evaluation and interaction state. As shown in Fig.~\ref{fig:p_action}, \textsc{Accept} is selected when all target labels are sufficiently reliable, \textsc{Edit} when the retained prediction supports localized correction, and \textsc{Generate} when the current result remains globally unreliable or further local refinement is unlikely to succeed. When \textsc{Edit} is selected, the controller additionally identifies the residual error to be targeted in the next iteration.

\subsection{Additional Analyses for Observations}
\label{sec:obs_supp}

\begin{figure*}[t]
  \centering
  \includegraphics[width=1\textwidth]{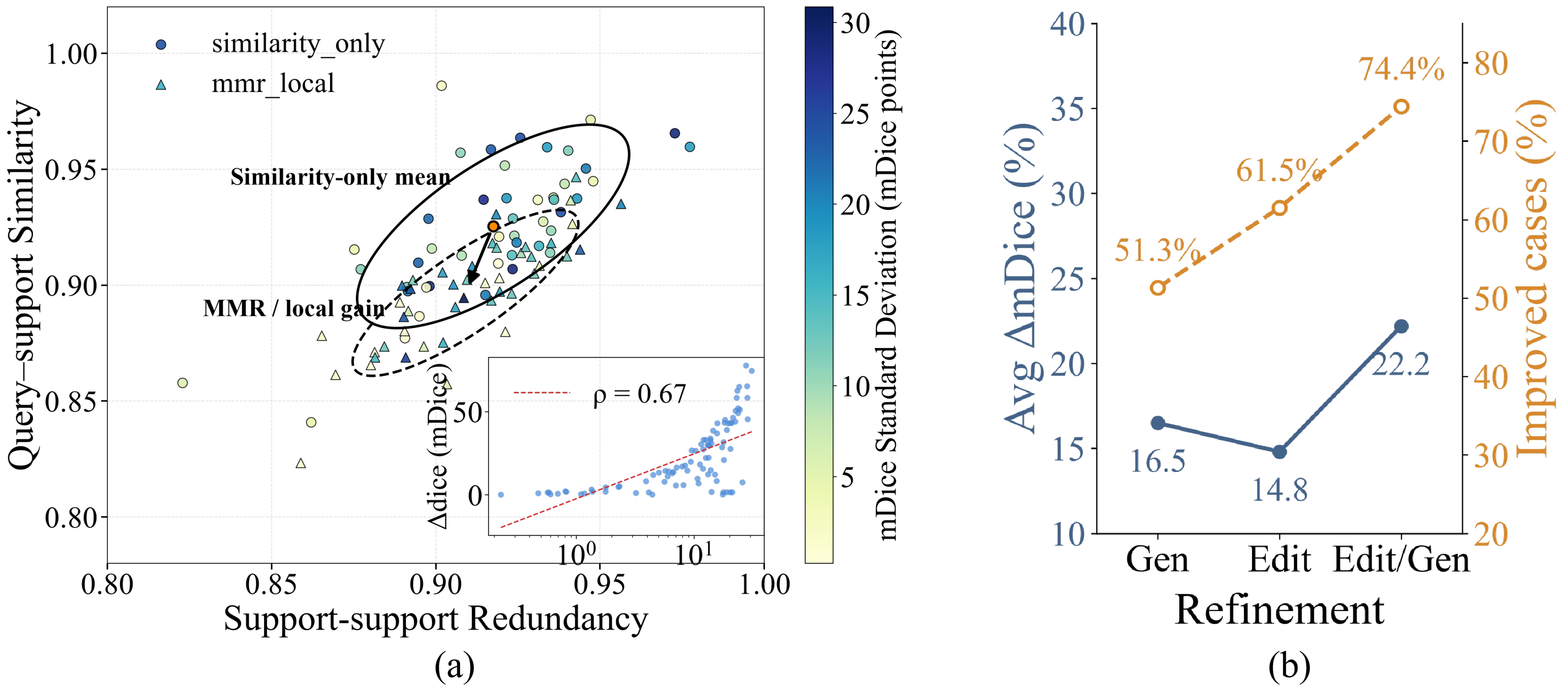}
  \caption{
Additional analyses for the observations in the main text.
(a) Diversity-aware retrieval broadens the accessible hypothesis space:
query--support similarity and support redundancy are compared between
similarity-only and MMR-based retrieval; marker color indicates hypothesis
dispersion measured by the standard deviation of Dice scores across sampled
predictions, and the inset shows its correlation with Best-of-4 improvement.
(b) Editing and resampling provide complementary refinement mechanisms:
the blue curve reports the mean Dice improvement and the orange curve reports
the proportion of improved cases under regeneration, targeted editing, and
their combination.
}
  \label{fig:obs_supp}
\end{figure*}

To further support the observations in the main text, we provide two additional analyses on inference-time exploration and refinement, as summarized in Fig.~\ref{fig:obs_supp}.
Fig.~\ref{fig:obs_supp}(a) analyzes diversity-aware retrieval and shows that varying retrieved supports further broadens the accessible hypothesis space.
Fig.~\ref{fig:obs_supp}(b) examines the complementary roles of editing and resampling in prediction refinement.

\paragraph{Diverse retrieval further broadens the accessible hypothesis space.}
Repeated sampling explores hypotheses within a fixed conditional context, while the context itself provides an additional source of diversity.
We therefore diversify inference by conditioning generation on different retrieved supports $S_i$, such that $M_i \sim p_{\theta}(M \mid I, T, S_i)$.
As shown in Fig.~\ref{fig:obs_supp}(a), under the same generation budget, distributing generations across diverse supports yields stronger Best-of-$N$ performance than repeatedly sampling from a fixed support.

We further investigate how to retrieve complementary supports by comparing similarity-based retrieval with Maximal Marginal Relevance (MMR).
MMR jointly considers query relevance and support redundancy, selecting the next support according to
\begin{equation}
S_i^* =
\arg\max_{S \in \mathcal{D}_s \setminus \mathcal{S}}
\left[
\lambda\,\operatorname{sim}(I,S)
-
(1-\lambda)\max_{S' \in \mathcal{S}}
\operatorname{sim}(S,S')
\right],
\end{equation}
where $\mathcal{S}$ denotes the already selected supports and $\lambda$ controls the trade-off between relevance and diversity.
Fig.~\ref{fig:obs_supp}(a) shows that similarity-only retrieval tends to achieve slightly higher query relevance but also higher mutual redundancy, whereas MMR produces supports that are less redundant while remaining highly relevant.
Specifically, MMR reduces the average support redundancy by $0.027$ while sacrificing only $0.009$ in query--support similarity, and improves the average Top-4 mDice from $60.04\%$ to $62.26\%$.

To further quantify hypothesis diversity, we measure the standard deviation
of Dice scores across sampled predictions. As shown in the inset of
Fig.~\ref{fig:obs_supp}(a), hypothesis dispersion is positively correlated
with Best-of-4 improvement ($\rho=0.67$), suggesting that broader hypothesis
exploration increases the chance of uncovering stronger predictions.
Overall, diversity-aware retrieval provides relevant yet less redundant contexts, enabling more effective exploration of the hypothesis space under a fixed inference budget.

\paragraph{Editing and resampling offer complementary mechanisms for prediction refinement.}
Editing exploits an existing promising hypothesis through targeted correction, but is less effective for fundamental prediction errors, whereas resampling explores alternative hypotheses without explicitly preserving useful structures in the current prediction.
We therefore investigate whether combining targeted editing with rollback and resampling can improve refinement robustness.

As shown in Fig.~\ref{fig:obs_supp}(b), generation alone yields an average
improvement of $16.5$ mDice points and improves $51.3\%$ of cases.
Targeted editing yields an average improvement of $14.8$ points and improves
$61.5\%$ of cases.
Combining editing with regeneration further increases the average improvement
to $22.2$ points and the proportion of improved cases to $74.4\%$.
These results demonstrate the complementary roles of editing and resampling:
editing enables targeted refinement of an existing hypothesis, while
regeneration expands the hypothesis space when local correction is insufficient.
Their combination provides the strongest overall refinement.

\subsection{Agent Behavior and Inference Efficiency}
\label{sec:efficiency}

\begin{table}[t]
\centering
\caption{
Average inference cost of SegBanana across datasets.
We report the average number of inference iterations, Nano Banana calls,
and retrieval calls per sample.
}
\label{tab:efficiency_cost}
\small
\setlength{\tabcolsep}{7pt}
\begin{tabular}{lccc}
\toprule
\textbf{Dataset}
& \textbf{Iter.}
& \textbf{Nano Banana Calls}
& \textbf{Ret. Calls} \\
\midrule
ACDC       & 2.47 & 4.52 & 3.88 \\
BraTS      & 1.76 & 3.40 & 2.26 \\
BUS-UCLM   & 1.60 & 3.18 & 1.78 \\
Drishti-GS & 2.40 & 4.76 & 3.28 \\
ISIC       & 1.70 & 3.24 & 1.52 \\
Kvasir     & 1.56 & 3.12 & 0.92 \\
RAVIR      & 2.25 & 4.50 & 3.33 \\
TNBC       & 2.36 & 4.72 & 2.72 \\
\midrule
\textbf{Overall}
& \textbf{1.94}
& \textbf{3.75}
& \textbf{2.32} \\
\bottomrule
\end{tabular}
\end{table}

\begin{table*}[t]
\centering
\caption{
Adaptive stopping, action selection, and retrieval behavior of SegBanana.
Stop@$k$ denotes the percentage of samples terminating after iteration $k$.
Action columns report the distribution of controller decisions after capping
\textsc{Generate} to at most two actions per iteration, and Retrieved denotes
the percentage of samples invoking support retrieval at least once.
}
\label{tab:stopping_behavior}
\small
\setlength{\tabcolsep}{5pt}
\begin{tabular}{lccc|ccc|c}
\toprule
\textbf{Dataset}
& \textbf{Stop@1}
& \textbf{Stop@2}
& \textbf{Stop@3}
& \textbf{Generate}
& \textbf{Edit}
& \textbf{Accept}
& \textbf{Retrieved} \\
\midrule
ACDC
& 15.0\% & 23.3\% & 61.7\%
& 76.6\% & 14.4\% & 9.0\%
& 100.0\% \\

BraTS
& 58.0\% & 8.0\% & 34.0\%
& 63.5\% & 15.3\% & 21.2\%
& 96.0\% \\

BUS-UCLM
& 66.0\% & 8.0\% & 26.0\%
& 55.8\% & 17.6\% & 26.7\%
& 90.0\% \\

Drishti-GS
& 28.0\% & 4.0\% & 68.0\%
& 65.3\% & 24.8\% & 9.9\%
& 100.0\% \\

ISIC
& 58.0\% & 14.0\% & 28.0\%
& 56.5\% & 20.6\% & 22.9\%
& 62.0\% \\

Kvasir
& 64.0\% & 16.0\% & 20.0\%
& 52.5\% & 20.3\% & 27.2\%
& 44.0\% \\

RAVIR
& 33.3\% & 8.3\% & 58.3\%
& 67.3\% & 16.4\% & 16.4\%
& 100.0\% \\

TNBC
& 20.0\% & 24.0\% & 56.0\%
& 57.1\% & 32.8\% & 10.1\%
& 100.0\% \\
\midrule
\textbf{Overall}
& \textbf{46.0\%}
& \textbf{14.0\%}
& \textbf{40.1\%}
& --
& --
& --
& \textbf{83.2\%} \\
\bottomrule
\end{tabular}
\end{table*}

Although SegBanana allows up to three inference iterations, its actual
computation is dynamically determined by the evolving prediction state.
We therefore analyze inference efficiency from two complementary perspectives:
the average computational cost and the adaptive termination, action selection,
and retrieval behavior of the controller.

\paragraph{Overall Inference Cost.}
Tab.~\ref{tab:efficiency_cost} reports the average inference cost across
datasets. Overall, SegBanana requires only $1.94$ inference iterations,
$3.75$ valid Nano Banana calls, and $2.32$ retrieval calls per sample.
Here, valid Nano Banana calls count successful generation outputs, with at
most two counted per iteration, matching the inference-time generation budget.

The inference cost varies substantially across datasets. Kvasir and BUS-UCLM
require only $1.56$ and $1.60$ iterations on average, whereas ACDC and
Drishti-GS require $2.47$ and $2.40$, respectively. A similar pattern appears
in visual-model usage: Kvasir requires only $3.12$ valid Nano Banana calls per
sample, while Drishti-GS and TNBC require $4.76$ and $4.72$. This variation
shows that SegBanana does not follow a fixed execution trajectory, but allocates
additional inference when the current prediction remains difficult to resolve.

\paragraph{Adaptive Termination.}
Tab.~\ref{tab:stopping_behavior} further summarizes the termination behavior.
Overall, $46.0\%$ of cases terminate after the first iteration and $14.0\%$
after the second, while $40.1\%$ reach the third iteration. Thus, $59.9\%$
of cases terminate before exhausting the maximum inference budget.

The stopping distribution is strongly task dependent. Most BUS-UCLM,
Kvasir, BraTS, and ISIC cases terminate after the first iteration, with
Stop@1 rates of $66.0\%$, $64.0\%$, $58.0\%$, and $58.0\%$, respectively.
In contrast, Drishti-GS, ACDC, RAVIR, and TNBC more frequently require the
full three-iteration budget, reaching Stop@3 rates of $68.0\%$, $61.7\%$,
$58.3\%$, and $56.0\%$. These results indicate that the controller adapts
not only the operation performed within each iteration but also the overall
inference depth.

\paragraph{Adaptive Action and Retrieval Usage.}
Generation remains the dominant action across all datasets, accounting for
$52.5\%$--$76.6\%$ of controller decisions. Editing is used more selectively,
ranging from $14.4\%$ on ACDC to $32.8\%$ on TNBC, consistent with its role
in locally refining predictions that already provide a reliable basis for
correction. The proportion of \textsc{Accept} decisions also varies markedly
across tasks, reflecting different stopping behaviors.

Knowledge retrieval is similarly adaptive rather than uniformly applied.
Overall, $83.2\%$ of cases invoke retrieval at least once. Retrieval is used
for all cases on ACDC, Drishti-GS, RAVIR, and TNBC, but only $44.0\%$ of
Kvasir and $62.0\%$ of ISIC cases require external support. This variation
shows that the controller acquires additional anatomical context according
to the task and evolving interaction state rather than enforcing retrieval
for every sample.

\subsection{Additional Case Studies}
\label{sec:additional_cases}

\begin{figure}[tb]
  \centering
  \includegraphics[width=1\textwidth]{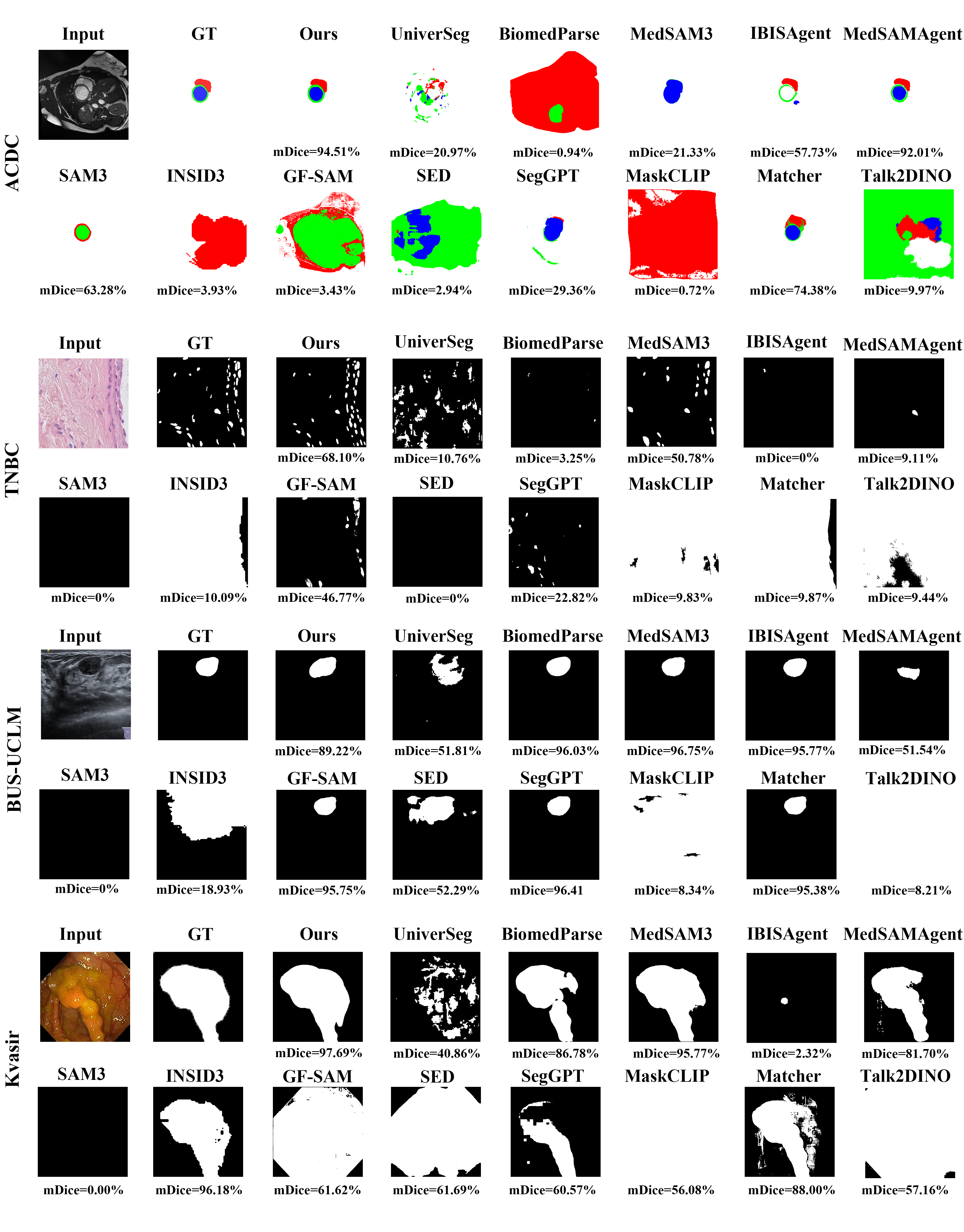}
  \caption{
  Additional qualitative comparisons on ACDC, TNBC, BUS-UCLM, and Kvasir.
  We compare SegBanana with representative medical-specific and generalist segmentation methods across cardiac MRI, histopathology, ultrasound, and endoscopy.
  }
  \label{fig:case1}
\end{figure}

\begin{figure}[tb]
  \centering
  \includegraphics[width=1\textwidth]{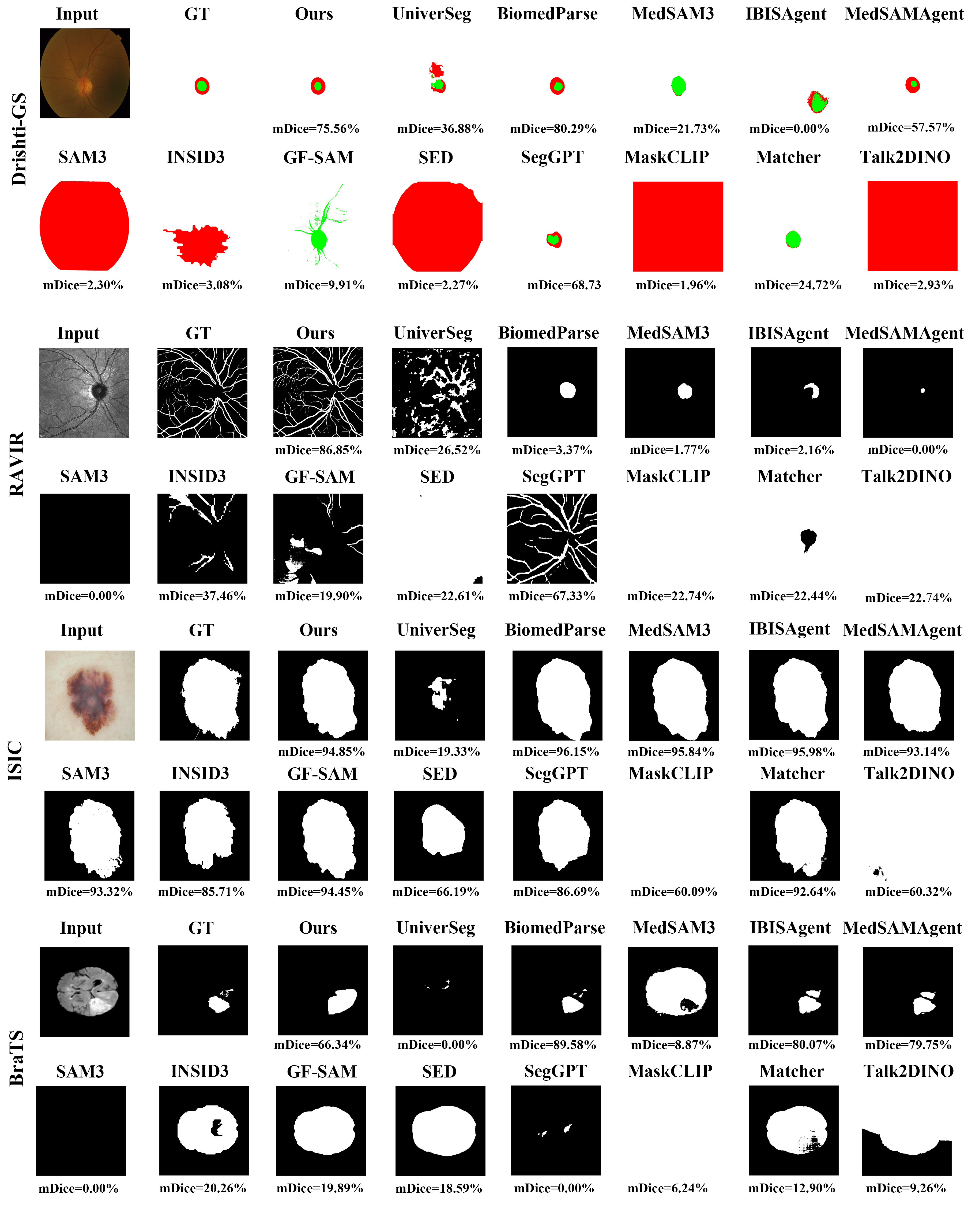}
  \caption{
  Additional qualitative comparisons on Drishti-GS, RAVIR, ISIC, and BraTS.
  The examples cover nested anatomical structures, fine vascular structures, skin lesions, and brain tumors, illustrating segmentation behavior across diverse targets and imaging modalities.
  }
  \label{fig:case2}
\end{figure}

Figs.~\ref{fig:case1} and~\ref{fig:case2} provide additional qualitative comparisons across eight datasets.
SegBanana produces predictions that remain visually consistent with the target anatomy across substantially different modalities and structural characteristics.
For relatively compact targets such as breast lesions, polyps, skin lesions, and brain tumors, its predictions generally preserve the overall target extent and shape, whereas several baselines exhibit substantial under-segmentation, over-segmentation, or foreground--background confusion.
For more structurally demanding tasks, the differences become more apparent: SegBanana better preserves the nested disc--cup relationship in Drishti-GS and the thin, branching vascular structures in RAVIR, while several competing methods collapse these structures into coarse regions or miss large portions of the target.

The histopathology and cardiac examples further illustrate the challenge of transferring across domains.
On TNBC, several methods produce sparse, fragmented, or nearly empty masks, whereas SegBanana captures a denser distribution of nuclei consistent with the reference.
On ACDC, SegBanana better preserves the relative spatial arrangement of multiple cardiac structures, while some baselines confuse labels or substantially distort their extent.
Overall, these examples qualitatively support the broad transferability of SegBanana across targets ranging from compact lesions to fine and multi-structure anatomy.

\end{document}